\documentclass[journal]{IEEEtran}
\ifCLASSINFOpdf
  \usepackage[pdftex]{graphicx}
\else
\fi
\usepackage[cmex10]{amsmath}
\usepackage{amsfonts}
\usepackage{amssymb}\usepackage{amsthm}\usepackage{bm}

\usepackage{amsfonts}

\usepackage{algorithmic}
\usepackage{algorithm}
\usepackage{boxedminipage}

\usepackage{xcolor}

\usepackage[normalem]{ulem}
\usepackage{verbatim}

\begin{document}
\title{Efficient Architecture for Deep Neural Networks with Heterogeneous Sensitivity}

\author{Hyunjoong~Cho,
	Jinhyeok~Jang,
	Chanhyeok~Lee,
        and~Seungjoon~Yang,~\IEEEmembership{Member,~IEEE}% <-this % stops a space
\thanks{
H. Cho, C. Lee, and S. Yang are with School of Electrical and Computer Engineering, Ulsan National Institute of Science and Technology
(UNIST), Ulsan, Korea (e-mail: hyeunjoong@unist.ac.kr, brandon27@unist.ac.kr, syang@unist.ac.kr). 
}
\thanks{J. Jang is with Electronics and Telecommunications Research Institute (ETRI), Daejeon, Korea (e-mail:jangjh6297@gmail.com).}
\thanks{
This work was supported by National Research Foundation of Korea under Grant NRF-2016R1D1A1B01016041.}
\thanks{
This work was supported by the Ulsan National Institute of Science and Technology Free Innovation Research Fund under Grant 1.170067.01
}
}


\markboth{Journal of \LaTeX\ Class Files,~Vol.~14, No.~8, August~2015}%
{Shell \MakeLowercase{\textit{et al.}}: Bare Demo of IEEEtran.cls for IEEE Journals}



\maketitle

\begin{abstract}
This work presents a neural network that consists of nodes with heterogeneous sensitivity. Each node in a network is assigned a variable that determines the sensitivity with which it learns to perform a given task. The network is trained by a constrained optimization that maximizes the sparsity of the sensitivity variables while ensuring the network's performance. As a result, the network learns to perform a given task using only a small number of sensitive nodes. Insensitive nodes, the nodes with zero sensitivity, can be removed from a trained network to obtain a computationally efficient network. Removing zero-sensitivity nodes has no effect on the network's performance because the network has already been trained to perform the task without them. The regularization parameter used to solve the optimization problem is found simultaneously during the training of networks. To validate our approach, we design networks with computationally efficient architectures for various tasks such as autoregression, object recognition, facial expression recognition, and object detection using various datasets. In our experiments, the networks designed by the proposed method provide the same or higher performance but with far less computational complexity.
\end{abstract}

\ifCLASSOPTIONpeerreview
\begin{center} \bfseries EDICS Category: 3-BBND \end{center}
\fi
\IEEEpeerreviewmaketitle

\section{Introduction}

\IEEEPARstart{N}{eural} networks often consist of a large number of nodes arranged in deep layers. As researchers have addressed an increasingly wide variety of problems using neural networks, these network architectures have become increasingly complicated. High-performance networks often contain thousands of nodes \cite{krizhevsky2012imagenet, szegedy2013intriguing, simonyan2014very, he2016deep, redmon2016you}. One property of neural networks is that nodes share the workload \cite{baldi1989neural}. A network is trained to perform a given task utilizing all the available nodes. Thus, all the nodes provided in a network architecture contribute towards solving a given problem. 

% pruning
As neural networks are implemented on platforms with less computational power, and as neural networks are asked to perform more tasks on platforms with massive but still finite computational power, designing a network with less computational complexity has become an important issue \cite{reed1993pruning, cheng2017survey}. Numerous studies have been conducted to determine how to prune network nodes and obtain a computationally efficient network  \cite{mozer1989skeletonization, lecun1990optimal, karmin1990simple, hassibi1993second, han2015learning, ishikawa1996structural, collins2014memory, li2016pruning, zhou2016less, jang2018deep}. In these approaches, the networks are first trained with a large amount of nodes; then, the importance of each node is evaluated by analyzing the node weight using various measures. Finally, the nodes with less importance are removed from the trained network. However, because all the nodes in trained network share the workload, removing a node from a trained network---even if its node weight indicates that is has lesser importance, will degrade the network's performance. Consequently, pruning approaches are usually followed by retraining the pruned network to recover the performance losses from pruning.  

% proposed
In this paper, we propose a network that consists of nodes with heterogeneous sensitivity. Each node in a network is assigned a variable that determines its sensitivity to learn a given task. Then, the network learns to perform the task by relying more on the sensitive nodes and less on the insensitive nodes. In extreme cases where the sensitivity of a node is zero, the network does not utilize the node at all; it is essentially disconnected. Node sensitivity is learned during training through a constrained optimization. The sparsity of the sensitivity is maximized while the network performance is constrained within a certain range. As a result, the network learns to perform a given task using only a small number of sensitive nodes. By simply removing the nodes with zero sensitivity, the computationally efficient architecture of a deep network for a given task is obtained. The regularization parameter used for the constrained optimization is determined simultaneously during the training based on the L-curve, which has previously been used  to determine the regularization parameters for inverse problems \cite{hansen1993use, hansen1999curve, hansen2005rank, hansen2006deblurring}.

% implementation
We assign sensitivity to nodes in a network by introducing a layer we call a sensitivity layer. The sensitivity layer can be implemented as a special type of dense or convolutional layer. Then, a network that includes these new sensitivity layers can be implemented and trained using functions available in standard deep learning packages \cite{jia2014caffe, chollet2015keras, abadi2016tensorflow}. Our approach does not require any special optimization routine to solve the constrained optimization problem that we designed to enforce the sparsity of the sensitivity variables while training.

% experiments & discussion
We designed computationally efficient architectures with heterogeneous sensitivity for various tasks such as autoregression, object recognition, facial expression recognition, and object detection using various datasets. We first applied networks with heterogeneous sensitivity to design a simple autoencoder and a deep convolutional neural network (CNN) for analysis. The effects of the regularization parameters on the network's performance and architecture are analyzed through the L-curve. Simultaneous selection of the regularization parameter during the training using the proposed algorithm is validated. Then, we applied the sensitivity layers to deep networks to solve various problems. Experiments are performed using various networks, autoencoder, CNN, LeNET \cite{krizhevsky2012imagenet}, VGG \cite{krizhevsky2012imagenet}, ResNet \cite{he2016deep}, and YOLO  \cite{redmon2016you}, and various dataset, Gaussian, MNIST  \cite{lecun1998gradient}, CIFAR-10 \cite{krizhevsky2009learning}, CK+ \cite{lucey2010extended}, VOC  \cite{Everingham15}, and ImageNet \cite{deng2009imagenet} for various types of classifications. By introducing nodes with heterogeneous sensitivity to the networks and enforcing the sparsity of the sensitivity, we were able to design networks that consist of notably fewer nodes but that exhibit the same or even better performance. The proposed method can be used to design an efficient network containing the optimal number of nodes.

% layout
The rest of this paper is organized as follows. We introduce networks with heterogeneous sensitivity and the constrained optimization to determine efficient network architecture in Section \ref{sec:network} and \ref{sec:optimal}, respectively. An algorithm for simultaneous selection of the regularization parameter during the network training is presented in Section \ref{sec:lcurve}. The implementation of the nodes with heterogeneous sensitivity as the sensitivity layer is discussed in Section \ref{sec:implementation}. Comparisons to pruning approaches to find efficient network architecture are given in Section \ref{sec:comparison}. We present the experimental results and discussions for the autoencoder in Section \ref{ex:gaussian} and \ref{ex:ae}. We find efficient network architecture of deep networks with heterogeneous sensitivity in Section \ref{ex:cnncifar} to \ref{ex:yolovoc}. We compare the performance and complexity of the optimal deep networks to those of pruned networks reported in literatures in Section \ref{ex:lenetmnist} to \ref{ex:resnetimagenet}. Section \ref{sec:conclusion} concludes the paper.

\section{Efficient Architecture for Deep Neural Networks}
\label{sec:proposed}

\subsection{Neural Networks with Heterogeneous Sensitivity}
\label{sec:network}

\begin{figure}[!t]
	\centering		
	
	\begin{minipage}{0.75\linewidth}		
		\centering
		{\includegraphics[trim = 0 0 0 0, clip, width=\linewidth]{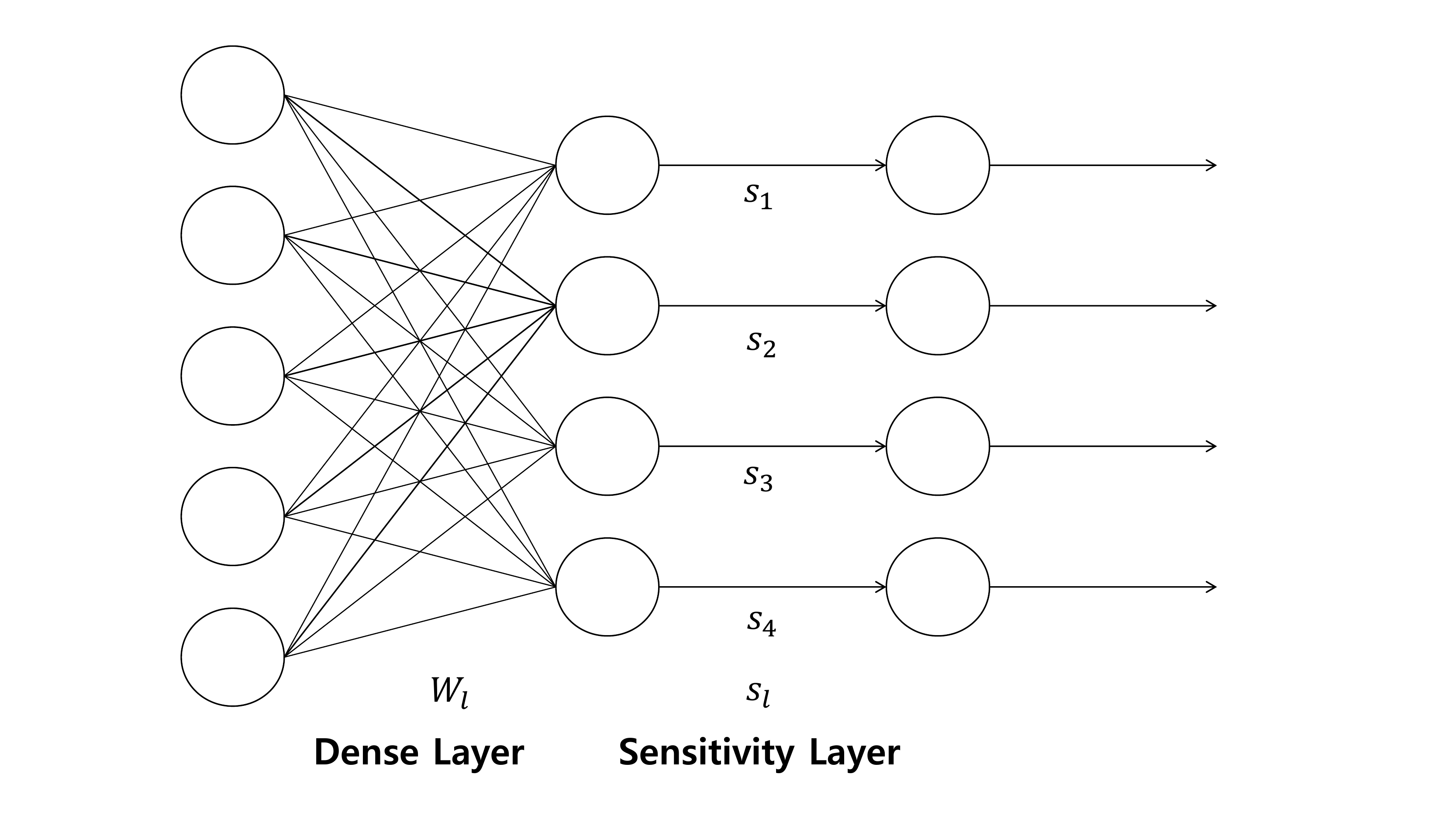}}%
		
		{\footnotesize (a)}
	\end{minipage}%
		
	\begin{minipage}{0.75\linewidth}		
		\centering
		{\includegraphics[trim = 0 0 0 0, clip, width=\linewidth]{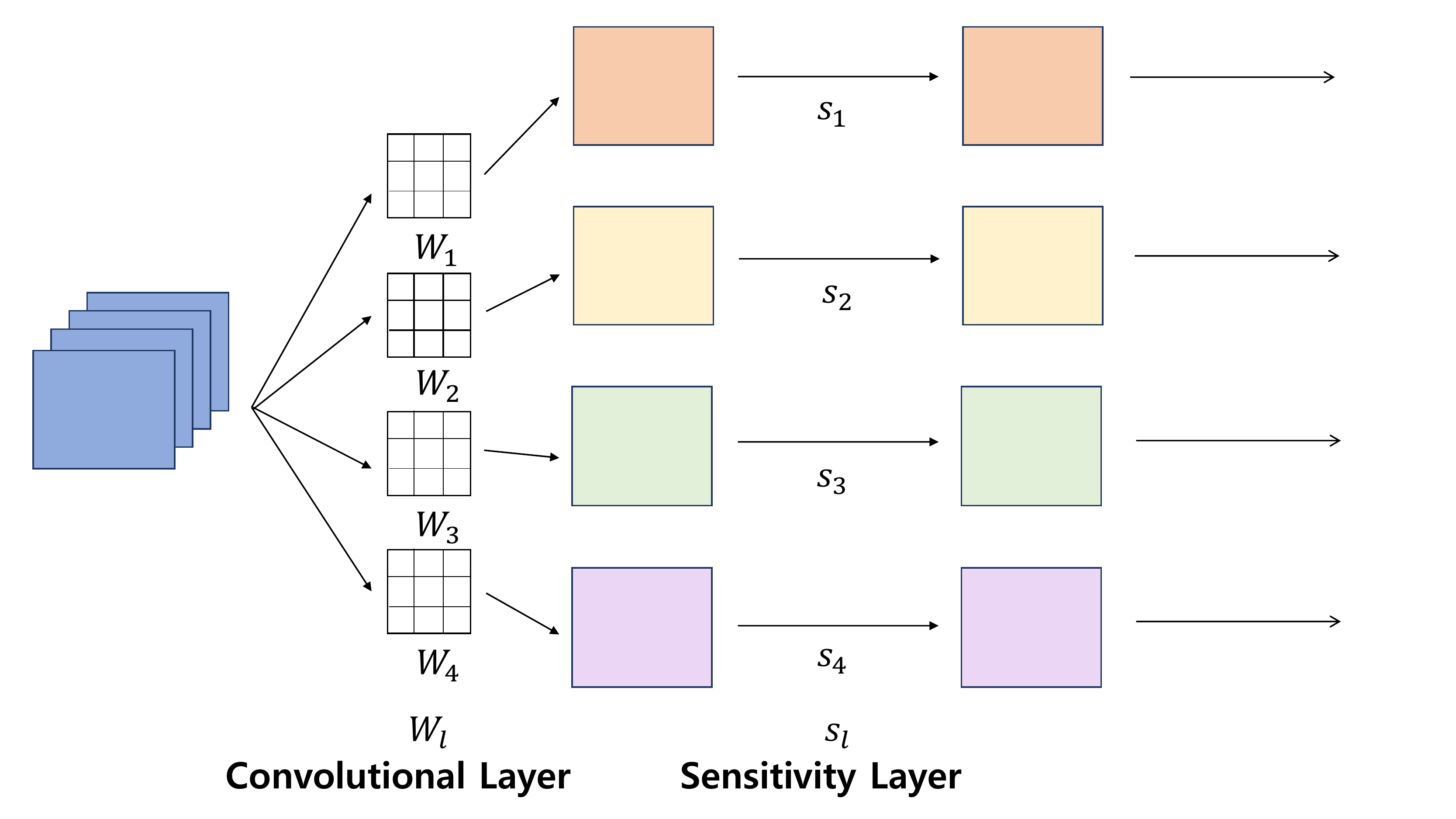}}%
		
		{\footnotesize (b)}
	\end{minipage}%
	\caption{A layer schematic in the proposed network with heterogeneous sensitivity: (a) dense layer; (b) convolutional layer.
	}
	\label{fig:schematics}
\end{figure}

% architecture
Consider a network whose $l$th layer consists of the following operations. The intermediate output $u_i^{l}$ is computed by either a dense layer,
\begin{equation}
	u_i^{l} = \sum_{j=1}^{n_{l-1}} W_{ij}^l x_j^{l-1},
	\label{eq:dense}
\end{equation}
or by a convolutional layer,
\begin{equation}
	u_i^{l} = \hbox{conv}(W_{i}^l, x^{l-1}),
	\label{eq:dense}
\end{equation}
where $u_i^l$ is the $i$th intermediate output node, $x_j^{l-1}$ and $x^{l-1}$ are the the $j$th nodes and the node volume in the $(l-1)$th layers, respectively. The network parameters $W_{ij}^l$ and $W_{i}^l$ denote a weight in the dense layer and a filter in the convolutional layer, respectively. The number of nodes in the $l$th layer are denoted as $n_l$. The intermediate output is activated by an activation function:
\begin{equation}
	v_i^{l} =  f^l(u_i^l).
	\label{eq:activation}
\end{equation}
The activated output $v_i^{l}$ is weighted by a newly introduced layer: 
\begin{equation} 
	x_i^{l} = s^l_i v_i^{l}, 
	\label{eq:sensitivity}
\end{equation}
where $s^l_i$ is a variable that determines the sensitivity of the $i$th node in the $l$th layer $x_i^l$. We denote this new layer as a sensitivity layer. The schematics of the proposed network architecture is given in Fig. \ref{fig:schematics}.

% fig schematics

% sensitivity
The sensitivity variable $s^l_i$ in the sensitivity layer allows us to apply heterogeneous sensitivity to the nodes in a network. Consider a network trained to minimize a cost function $E$ using  backpropagation \cite{rumelhart1986learning, haykin1994neural}. The weight matrix $\mathbf{W}_l$, whose element is the weight $W_{ij}^l$ of the $l$th layer, is updated by
\begin{equation}
	\mathbf{W}_l \gets \mathbf{W}_l - \eta \bm{\delta}_l \mathbf{x}_{l-1}^\mathsf{T},
	\label{eq:update}
\end{equation}
where $\eta$ is the step size and $\mathbf{x}_{l-1}$ is a vector whose elements are $x_i^{l-1}$. In the $l$th layer, each element of the sensitivity vector $\bm{\delta}_l$ is 
\begin{equation}
	\delta_i^{l} = s^l_i (\sum_{k=1}^{n_{l+1}} \delta_k^{l+1} W_{ki}^{l+1}) \frac{\partial f^l (u_i^{l})}{\partial u_i^l}.
	\label{eq:sensitivity2}
\end{equation}
The variable $s^l_i$ is a weight that reflects the sensitivity of a node. When the weights are updated, nodes with larger values of $|s^l_i|$ will respond more sensitively than those with smaller $|s^l_i|$ values. In the extreme case, when $s^l_i=0$, the node is completely insensitive. These zero-sensitivity nodes can be regarded as disconnected nodes.

\subsection{Optimization for Efficient Network Architecture}
\label{sec:optimal}

% training optimization
Node sensitivity (the sensitivity variable $s^l_i$) can be determined during the training so that a network learns to perform a given task using only a small number of sensitive nodes. To accomplish this, we designed an optimization problem to train the network: 
\begin{equation}
\begin{split}
	%\underset{\mathbf{s}^1,\mathbf{s}^2,\cdots, \mathbf{s}^L}{\textnormal{minimize}} \quad & \sum_{l=1}^L \|\mathbf{s}^l\|_1\\
	\textnormal{minimize} \quad & \sum_{l=1}^L \|\mathbf{s}_l\|_1\\
	\textnormal{subject to} \quad & E < \epsilon,
	%\\ & \mathbf{s}^l \geq \mathbf{0},
	\end{split}
	\label{eq:optimization1}
\end{equation}
where $\mathbf{s}_l$ is a vector whose elements are $s^l_i$ and $E$ is a deviation penalty that measures the deviations of network outputs from the ground truth values. The cost function, which is the sum of the $\ell_1$ norm of $\mathbf{s}_l$, makes the vector $\mathbf{s}_l$ sparse, so that the network uses only a few sensitive nodes and includes as many disconnected nodes as possible. The constraint on $E$ guarantees the performance of the network within the threshold $\epsilon$. The deviation penalty $E$ is a typical cost function used to train a network. The last layer of a network is usually determined by a specific task. We set the sensitivity variable for the last layer $s^L_i$ to one.

% optimization
The optimization problem can be rewritten as follows:
\begin{equation}
	\underset{\mathbf{W}_1,\mathbf{W}_2,\cdots, \mathbf{W}_L}{\underset{\mathbf{s}_1,\mathbf{s}_2,\cdots, \mathbf{s}_L}{\textnormal{minimize}}}
	\quad E + \lambda \sum_{l=1}^L \|\mathbf{s}_l\|_1,
	\label{eq:optimization2}
\end{equation}
where the regularization parameter $\lambda$ weighs the deviation penalty $E$ and the sparsity penalty $\sum \|\mathbf{s}_l\|_1$. When $\lambda$ is large, the sparsity penalty dominates the cost function in \eqref{eq:optimization2}. The trained network will consist of a small number of sensitive nodes with as many insensitive (i.e.,  disconnected) nodes as possible. However, because the deviation penalty $E$ is neglected during the training, the network will fail to provide accurate outputs. In contrast, when $\lambda$ is small, the deviation penalty dominates the cost function, and the network will be trained to provide accurate outputs but will utilize most of the available nodes. Thus few insensitive (i.e., disconnected) nodes will exist. Ideally, the goal is to find a $\lambda$ that balances the deviation and sparsity penalties.

\subsection{Regularization Parameter Selection Via L-Curve}
\label{sec:lcurve}

% L-curve
An L-curve is a plot of the two penalties for various values of $\lambda$. The L-curve has previously been used in inverse problems  \cite{hansen1993use, hansen1999curve, hansen2005rank, hansen2006deblurring}. In our problem, the L-curve shows the deviation penalty vs. the sparsity penalty for various values of $\lambda$. Let the deviation and sparsity penalties when the network is trained with the regularization parameter $\lambda$ be $E(\lambda)$ and $S(\lambda)$.
The L-curve is given by
\begin{equation}
	\{(\phi(E(\lambda)),\phi(S(\lambda))),\lambda > 0\},
	\label{eq:lcurve}
\end{equation}
where $\phi$ is an increasing function such as the log function. One region of the L-curve corresponds to solutions that are dominated by the deviation penalty, and another region of the L-curve corresponds to solutions that are dominated by the sparsity penalty. The curve generally has an L-shape. The corner of the L-curve provides a solution for which the two penalties are balanced. We use the L-curve to determine the best value of the regularization parameter $\lambda$ for the optimization problem in \eqref{eq:optimization2}.

% review
The operation of a deep network is usually composed of multiple  layers with non-linear activation function. Parameter selection methods based on diagonalization of a matrix that represents a linear operation of a system \cite{calvetti2000tikhonov, xie2009lanczos} are not applicable to determine the regularization parameter for deep networks. Also, the training of a deep network usually requires large amount of computation. Parameter selection methods that find the maximum curvature point of the L-curve \cite{hansen1993use} are not practical, because the L-curve has to be constructed for many values of $\lambda$.

% algortihm
We determine the regularization parameter $\lambda$ simultaneously during the training of a deep network. The network training is initialized with random weight $\mathbf{W}_l$'s and a small initial regularization parameter $\lambda_0$. The network is trained for a given number of epochs and the deviation cost $E$ is measured. While the measured deviation penalty is smaller than $\varepsilon$, the regularization parameter $\lambda$ is increased by $\Delta \lambda$, the sensitivity variables are initialized randomly, and the network is trained for a given number of epochs. When the regularization parameter $\lambda$ becomes too large, the network cannot provide outputs close to the ground truth. The deviation penalty $E$ increases substantially and becomes larger than $\varepsilon$. We find the largest regularization parameter $\lambda$ value that provides the deviation penalty $E\leq \varepsilon$. Then, the training of the network continues with the found regularization parameter until the termination condition for the network training is met. The algorithm for the simultaneous network training and regularization parameter selection is given in Algorithm \ref{alg:training}.

\begin{algorithm}[t!]
\caption{Simultaneous Training and Parameter Selection}
\label{alg:training}
%\begin{boxedminipage}{\linewidth}
	\begin{algorithmic}[1]
	\STATE $\lambda \gets \lambda_0$,$E\gets \varepsilon$.
	\STATE initialize $\mathbf{W}_l$'s.
	\WHILE{Termination condition is not met}
	\WHILE{$E \leq \varepsilon$}
		\STATE $\lambda \gets \lambda + \Delta \lambda$
		\STATE Initialize $\mathbf{s}_l$'s.
		\STATE Train the network for a number of epochs
		\STATE Measure $E$.
	\ENDWHILE
	\STATE Train the network.
	\ENDWHILE
	\end{algorithmic}
%\end{boxedminipage}
\end{algorithm}

\subsection{Sensitivity Layer Implementation}
\label{sec:implementation}

% implementation
The sensitivity layer can be regarded as a special type of dense or a convolutional layer. A sensitivity layer added after a dense layer with $n_l$ nodes can regarded as a set of $n_l$ (dense) layers with one input node, one output node, and one weight. A sensitivity layer added after a convolutional layer with $n_l$ nodes can be regarded as a set of $n_l$ convolutional layers with one input node, one output node, and a single one-by-one filter. Hence, the sensitivity layer can be implemented using layer definitions and functions already available in deep learning packages \cite{jia2014caffe, chollet2015keras, abadi2016tensorflow}. 

% training
Moreover training a network that includes sensitivity layers can be accomplished with various training methods already available in deep learning packages. The weights in the dense or convolutional layers are updated by
\begin{equation}
	W^l_{ij} \gets W^l_{ij} - \eta \frac{\partial E}{\partial W^l_{ij}}, 
\end{equation}
and the parameters in the sensitivity layers are updated by
\begin{equation}
	s^l_i \gets s^l_i - \eta  \frac{\partial E}{\partial s^l_i} - \eta \lambda \frac{s^l_i}{|s^l_i|}.
\end{equation}
These approaches can be implemented as the training for either dense or convolutional layers under $\ell_1$ regularization. Hence, by regarding the sensitivity layers as simply special cases of dense or convolutional layers, we can implement and train a network with the sensitivity layers using standard deep learning packages. Our approach does not require any special optimization routines to solve the constrained optimization problem.

\subsection{Comparison with Pruning Methods}
\label{sec:comparison}

% pruning approaches
Previous studies have investigated how to design compact and efficient networks to allow deep networks to be deployed on devices with restricted computational capabilities. A survey of efficient deep network design methods can be found in \cite{reed1993pruning, cheng2017survey}. Many approaches can prune nodes with high computational complexity and memory requirements to obtain a more efficient network. In these pruning approaches, a network is first trained, and then the importance of each node is evaluated with various measures. Finally, the less important nodes are pruned from the trained network. The issue of how the cost function used for the training changes with small weight perturbations was analyzed In \cite{mozer1989skeletonization, lecun1990optimal, karmin1990simple, hassibi1993second}. For example, the Hessian of a cost function provides information on how small weight changes affect the cost. Connections in a trained network that induce insignificant changes in the cost function were removed from the network. In \cite{han2015learning, ishikawa1996structural, collins2014memory, li2016pruning, zhou2016less}, the importance of weights in a trained network was evaluated using the $l_2$, $l_1$, and $l_0$ norms of the node weights or distances between the weights \cite{ayinde2018building}. Then, the less important connections between nodes were  removed from the trained network based on the evaluated measures. To encourage a network to have node weights that result in smaller measures, regularization by the $l_2$, $l_1$, and $l_0$ norms of the node weights is used during training. 
 
\begin{figure}[!t]
	\centering		
	
	\begin{minipage}{0.41\linewidth}		
		\centering
		{\includegraphics[trim = 0 0 300 0, clip, width=\linewidth]{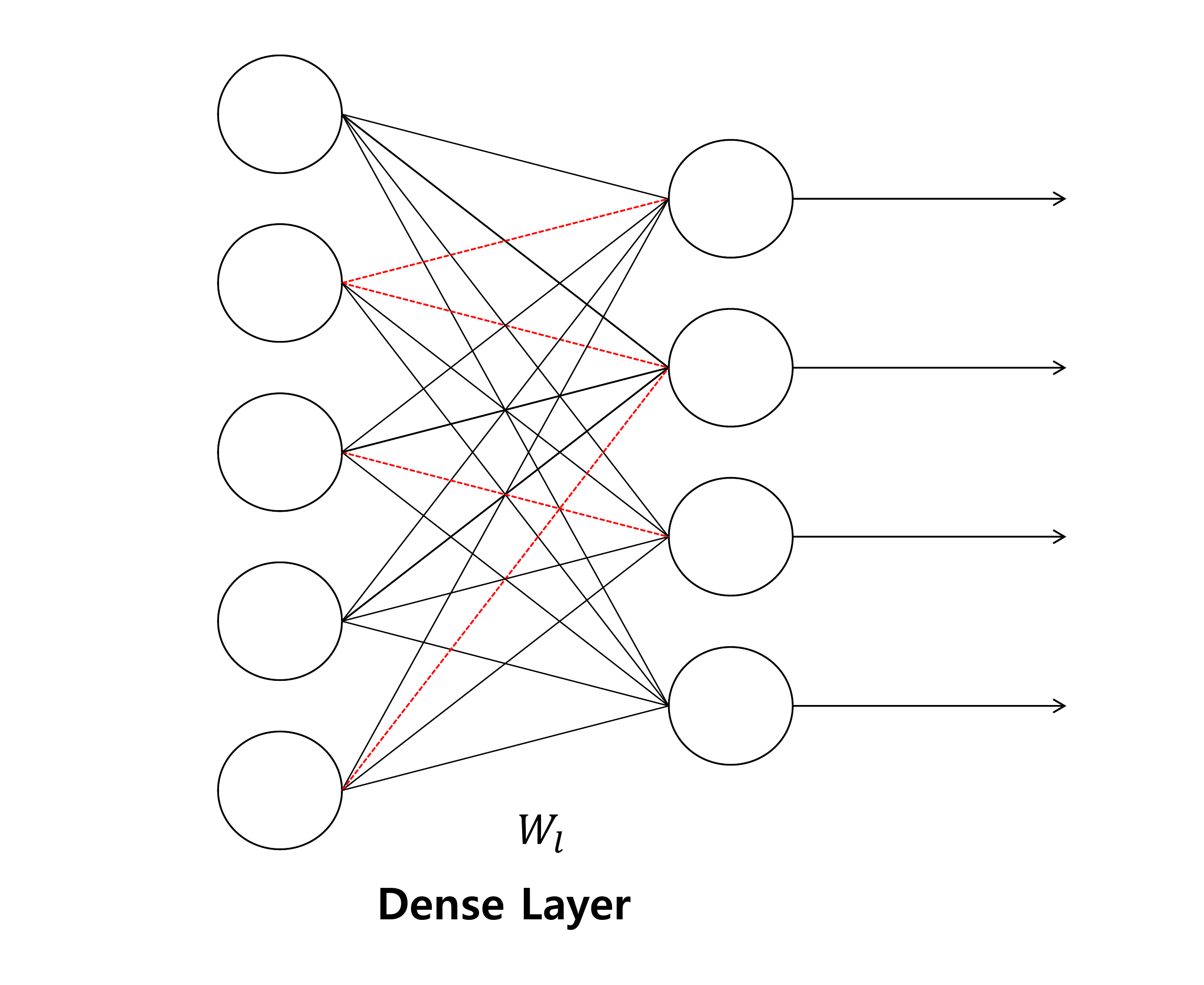}}%
		
		{\includegraphics[trim = 0 0 300 0, clip, width=\linewidth]{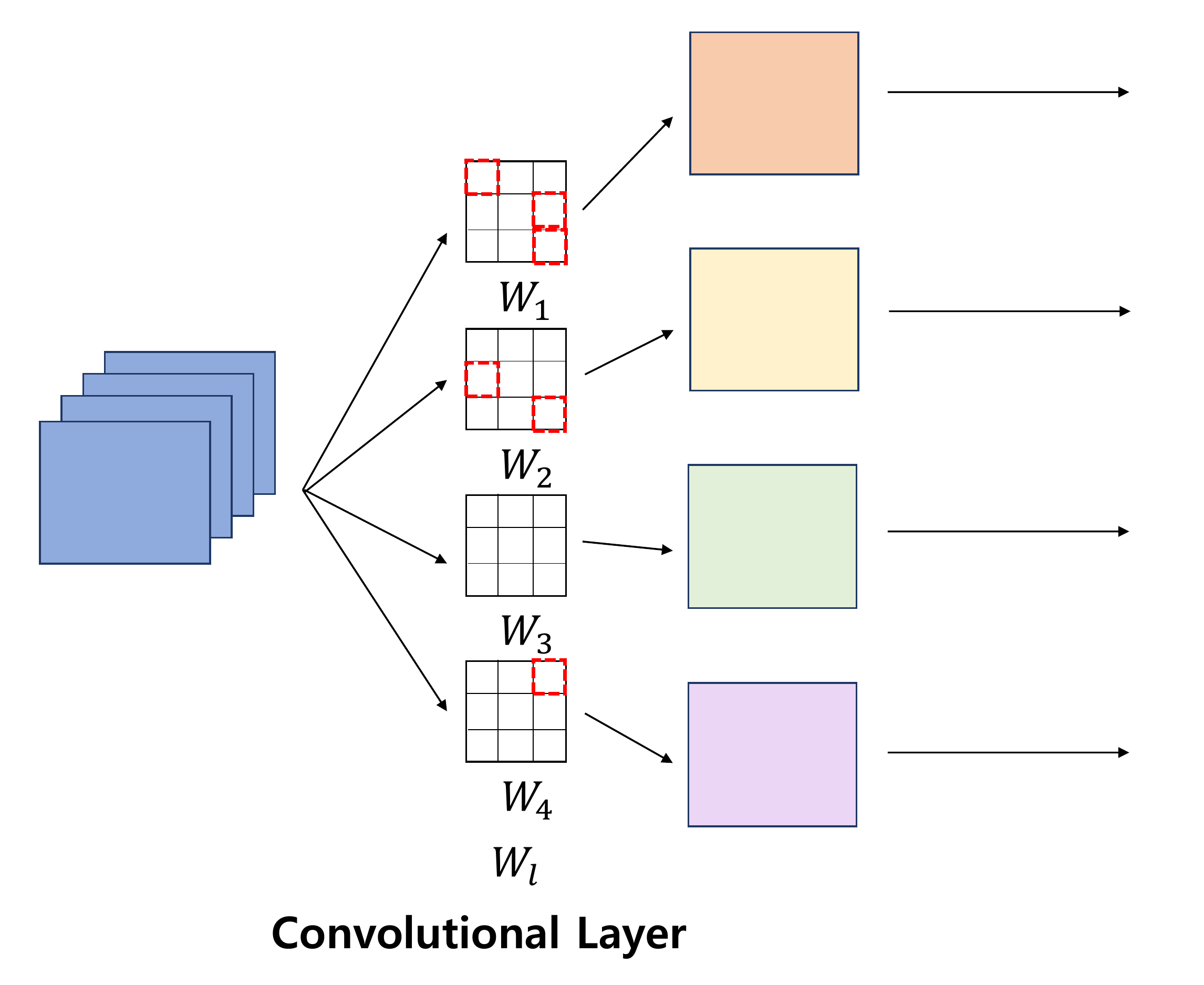}}%
		
		{\footnotesize (a)}
	\end{minipage}%
	\begin{minipage}{0.59\linewidth}		
		\centering
		{\includegraphics[trim = 0 0 0 0, clip, width=\linewidth]{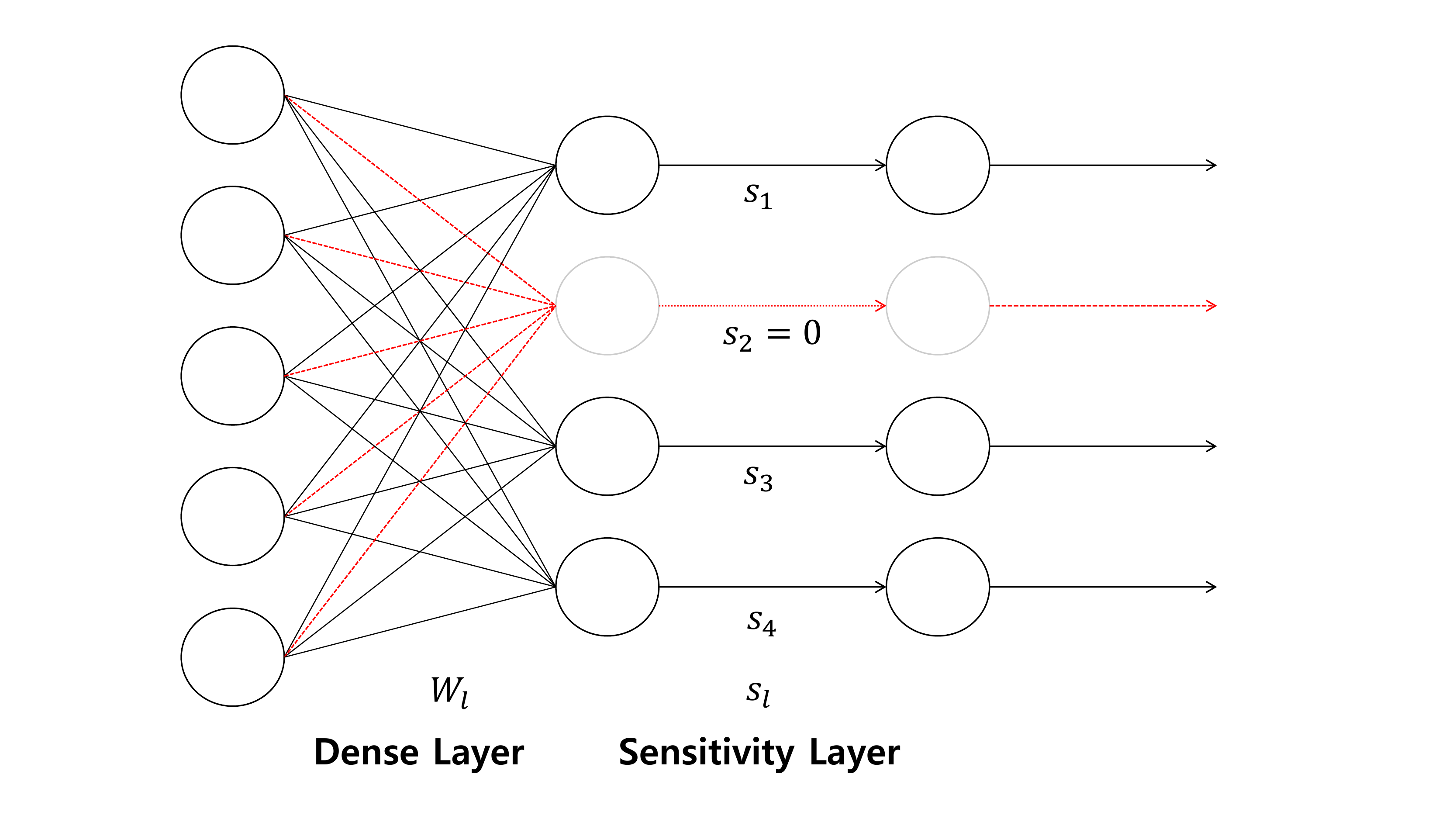}}%
		
		{\includegraphics[trim = 0 0 0 0, clip, width=\linewidth]{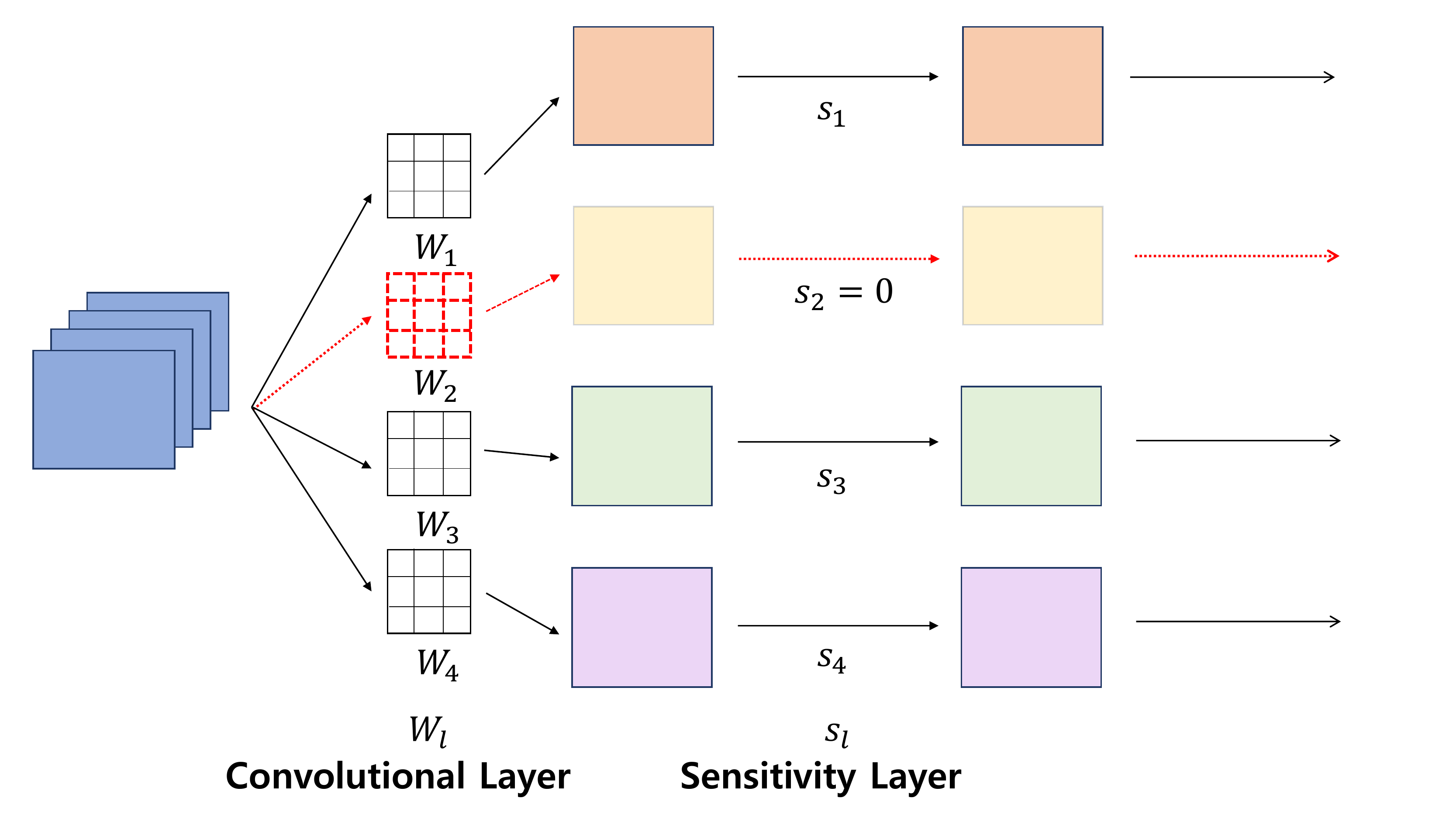}}%
		
		{\footnotesize (b)}
	\end{minipage}%
	\caption{Examples of removed connections in a network: (a) pruning approaches; (b) proposed method: 	top: dense layer; 	bottom: convolutional layer. Removed connections are indicated with red dotted lines.
	}
	\label{fig:schematicspruning}
\end{figure}

% comparison
Many pruning approaches use regularization as a function of node weights. As a results some connections in dense layers and some filter coefficients in convolutional layers have small values. Removing a node from a network entirely is not straightforward for dense layers and is difficult for convolutional layers. Examples of pruning with removed connections and filter coefficients are shown in Fig. \ref{fig:schematicspruning} (a), where prunned connection and prunned filter cofficients are denoted as red lines and blocks, respectively. The network is trained to perform a task utilizing all the available nodes. Because nodes in a network share the workload \cite{baldi1989neural}, removing a node---even one with a smaller measure of importance---from a trained network will degrade the network's performance. Consequently, pruning approaches typically retrain the pruned network to recover the performance loss from pruning.  

% ours
The proposed method uses regularization as a function of the sensitivity. As a result, removing zero-sensitivity nodes is straightforward because they are already disconnected by  the end of the training. Examples of disconnected nodes for dense and convolutional layers are shown in Fig. \ref{fig:schematicspruning} (b), where the nodes with zero sensitivity, denoted with red lines, can be removed from a trained network. 
The network is trained to perform a task utilizing only the sensitive nodes; therefore, removing zero-sensitivity nodes has no effect on the network's performance because the network has already been trained to perform the task without them.
% fig schematics

% comparison to jang
In \cite{jang2018deep}, networks with activation functions with nodewise variant slopes were introduced. Using this approach, the nodes with steeper slopes learn more important features, and vice versa. After training, the nodes with lower slopes are pruned from the trained network. The assignment of nodewise variant slopes to activation functions plays a role similar to that of the sensitivity layers presented in this paper; however, that study used a predefined set of values for the slopes regardless of the data. Because a predefined set of slopes does not reflect the actual data statistics, workload sharing still exists; hence, a performance loss occurs after pruning. Our proposed method can be viewed as an improvement of the work in \cite{jang2018deep} in which the slopes of the activation functions are learned from the data statistics by solving an optimization problem during the training. 

% comparison to he
In \cite{he2017channel}, a layer similar to the sensitivity layer in our proposed network was introduced to a trained network. The variables in the added layer were then used for pruning. An optimization problem was constructed to determine which nodes could be removed from the trained network while still ensuring the network's performance. Again, however, because nodes in a trained network share the workload, node pruning---even when done through an optimization approach, degrades the network's performance. In contrast, the variables in the sensitivity layers in our approach are found during network training, thus avoiding the need for further pruning. 

% comparison to wen
In \cite{wen2016learning}, a group lasso of node weights is used as a regularization factor. The group lasso enforces groupwise sparsity. By defining all the coefficients in a filter as a group, a node can be effectively removed from a convolutional layer. By defining only a part of the filter coefficients as a group, filters with different support levels can be used in a convolutional layer. Consequently, different network architectures can be designed by defining different groups. Such a design requires multiple regularization parameters; however, the study did not address how to choose the regularization parameters to obtain efficient architecture. In contrast, our approach uses regularization as a function of node sensitivity, which allows us to simply disconnect a node in both dense and convolutional layers. The regularization parameter is chosen using an L-curve to help find efficient architecture.

\section{Experiments and Discussions}
\label{sec:ex}

We first analyze a network with heterogeneous sensitivity using a simple autoencoder with Gaussian data and the MNIST dataset  \cite{lecun1998gradient} in Section \ref{ex:gaussian} and \ref{ex:ae}, respectively. The regularization parameter selection via the L-curve is discussed in Section \ref{sec:regularization} using the autoencoder with the MNIST dataset. Then, we find efficient network architecture of deep networks with heterogeneous sensitivity using a CNN with the CIFAR-10 dataset \cite{krizhevsky2009learning}, using VGG \cite{simonyan2014very} and ResNet \cite{he2016deep} with the CK+ dataset \cite{lucey2010extended}, and using YOLO \cite{redmon2016you} with the VOC dataset \cite{Everingham15} in Section \ref{ex:cnncifar} to \ref{ex:yolovoc}. We compare the performance and complexity of the proposed deep networks to those of pruned networks reported in literatures for LeNet \cite{krizhevsky2012imagenet} with the MNIST dataset, VGG and ResNet with the CIFAR-10 dataset, and ResNet with the ImageNet dataset \cite{deng2009imagenet} in Section \ref{ex:lenetmnist} to \ref{ex:resnetimagenet}.

\subsection{Autoencoder with Gaussian Data}
\label{ex:gaussian}

% set up
To understand how a network with sparse sensitivity variables is trained to perform a given task, consider a simple network in an autoregression setting. As an example, we use a network with two dense layers and linear activation. The sensitivity layer is implemented in the first layer. Then, network operation can be written as follows:
\begin{eqnarray}
	\mathbf{y} & = & \mathbf{A}\mathbf{x} \\ 
		& = & \mathbf{W}_2\mathbf{S}_1\mathbf{W}_1\mathbf{x} \\
		& = & \sum_{k=1}^{n_1} s^1_k \mathbf{w}^2_k (\mathbf{w}^1_k)^\mathsf{T} \mathbf{x},
		\label{eq:linearcomb}
\end{eqnarray}
where $\mathbf{W}_1$ and $\mathbf{W}_2$ are the weight matrices. The matrix $\mathbf{S}_1$ is a diagonal matrix whose diagonal elements are the sensitivity variables in $s^1_k$. The vectors $\mathbf{w}^2_k$ and $(\mathbf{w}^1_k)^\mathsf{T}$ are the $k$th column and row of $\mathbf{W}_2$ and $\mathbf{W}_1$, respectively. The network operation is written as a linear combination of $\mathbf{w}^2_k (\mathbf{w}^1_k)^\mathsf{T}$. The weights for the linear combination are given by the sensitivity variable $s^1_k$. The contribution of $\mathbf{w}^2_k (\mathbf{w}^1_k)^\mathsf{T}$ with small $s^1_k$ values to the operation of the network is small, and vice versa.

% pca
We find the columns of the matrices $\mathbf{W}_1$, $\mathbf{W}_2$ and the diagonal matrix $\mathbf{S}_1$ through the optimization problem in \eqref{eq:optimization2}. The sensitivity variable $\mathbf{s}_1$ obtained by the optimization will have many zero elements because sparsity is enforced. Without loss of generality, let the elements of $\mathbf{s}_1$ be sorted such that the $(K+1)$th to $n_1$th elements are zero. The contribution of $\mathbf{w}^2_k (\mathbf{w}^1_k)^\mathsf{T}$ with $s^1_k=0$ is zero. Hence, we can remove those terms from the linear combination in \eqref{eq:linearcomb}. Then, we approximate the network's operation by
\begin{equation}
	\mathbf{A} \approx \sum_{k=1}^K s^1_k \mathbf{w}^2_k (\mathbf{w}^1_k)^\mathsf{T}.
\end{equation}

\begin{figure}[t!]
	\centering		
	\begin{minipage}{0.5\linewidth}		
		\centering
		{\includegraphics[width=\linewidth]{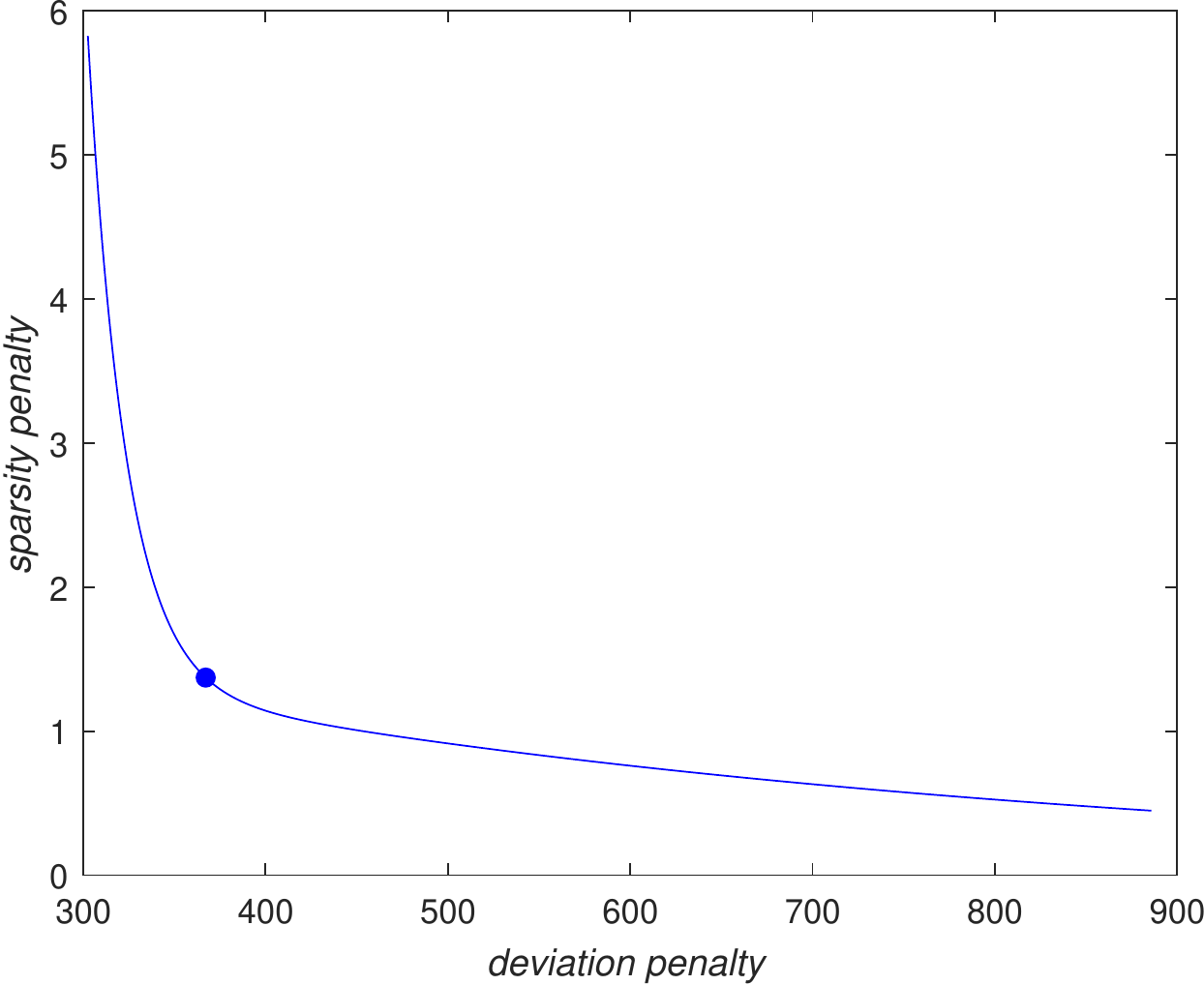}}%
	\end{minipage}%
	\begin{minipage}{0.5\linewidth}		
		\centering
		{\includegraphics[width=\linewidth]{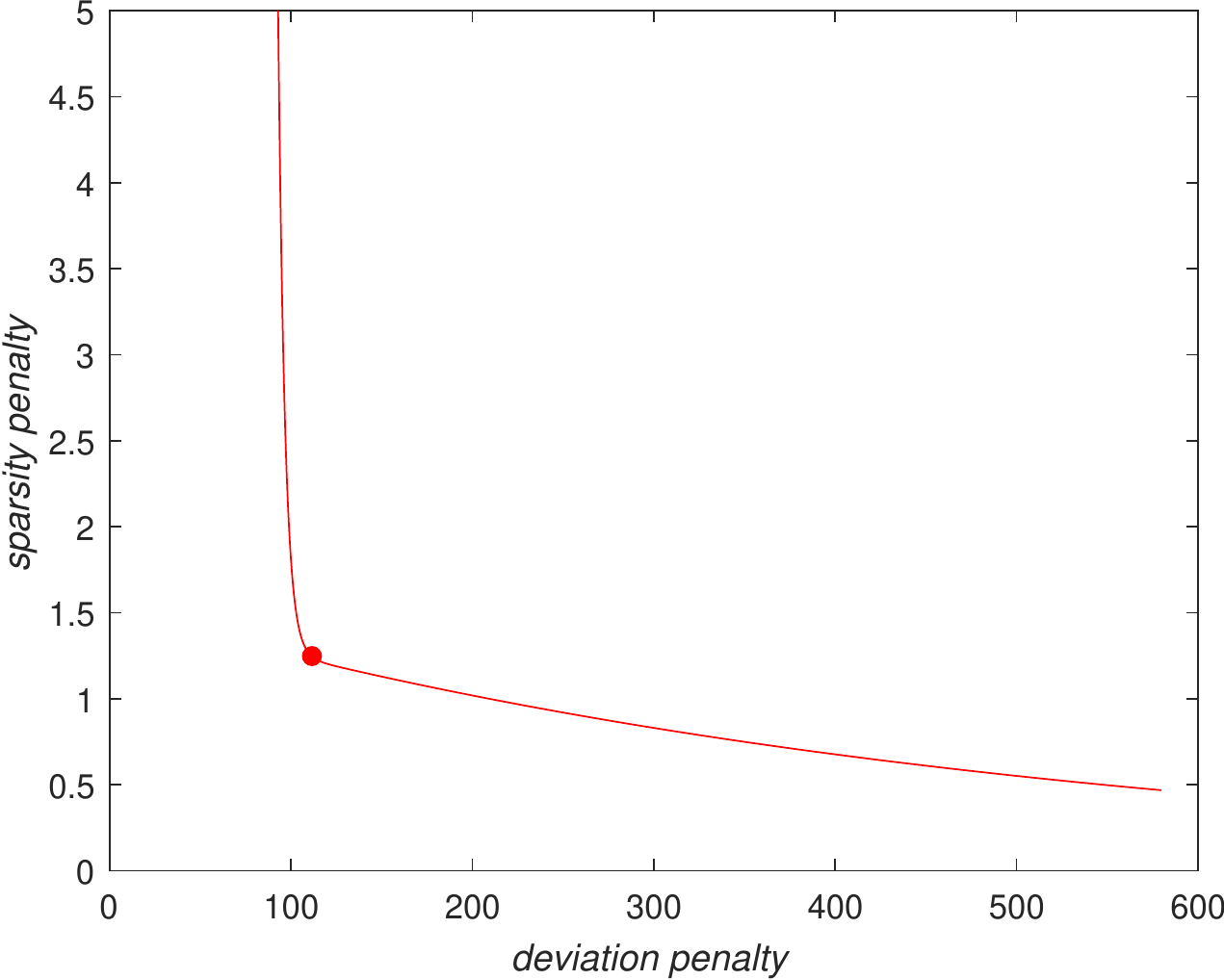}}%
	\end{minipage}%	
	
	\caption{Example of L-curve for autoencoder using Guassian dataset, showing $E$ vs. $\sum_{l=1}^L \|\mathbf{s}^l\|_1$ for different values of $\lambda$; 
	(a) For a case where 8 of 12 inputs are correlated with high variances and the number of hidden nodes is 12;
	(b) For a case where 4 of 12 inputs are correlated with high variances and the number of hidden nodes is 12.
	}
	\label{fig:lcurvegaussian}
\end{figure}

% experiment
Experiments are performed with 16, 12, and 8 inputs following zero mean Gaussian distribution. We consider cases where some inputs are correlated with large variances and the rest of the inputs are independent with small variances. The large and small variances are 1.0 and 0.0001, and the correlation coefficient is 0.9. The number of hidden nodes, $n_1$, are set to 16, 12, and 8. The autoencoders are trained by \eqref{eq:optimization2} for 50 times. Table \ref{tab:gaussian} shows the  average numbers of nodes with nonzero and zero sensitivity. The network is trained using the regularization parameter $\lambda$ corresponding to the corner point of the L-curves. Examples of the L-curves are shown in Fig. \ref{fig:lcurvegaussian}. The average number of sensitive nodes, with nonzero sensitivity, is close to the numbers of nodes with large variances. The proposed network with heterogeneous sensitivity trained via the optimization problem in \eqref{eq:optimization1} operates similarly to the principal component analysis \cite{jolliffe2011principal}---it represents the inputs through the small number of sensitive, or principal, nodes.

\begin{table}[t!]
	\centering
	\caption{Average Numbers of Sensitive Nodes Determined by Constrained Optimization using Gaussian Dataset}
	\label{tab:gaussian}
\begin{tabular}{cc|c|cc}										
\hline										
\# of inputs	&	\# of inputs	&	\# of hidden	&	\# of nodes	&	\# of nodes	\\	
with	&	with	&	nodes	&	with	&	with	\\	
$\sigma$=1.0	&	$\sigma=0.01$	&		&	$s \neq 0$	&	$s = 0$	\\	\hline
8	&	8	&	16	&	8.3	&	7.7	\\	
8	&	4	&	12	&	8.0	&	4.0	\\	
8	&	0	&	8	&	7.5	&	0.5	\\	\hline
4	&	12	&	16	&	4.0	&	12.0	\\	
4	&	8	&	12	&	4.1	&	7.9	\\	
4	&	4	&	8	&	4.1	&	3.9	\\	\hline
\end{tabular}																																																		
\end{table}

\subsection{Autoencoder with MNIST Dataset}
\label{ex:ae}

% set up
We analyze a simple network with heterogeneous sensitivity trained with a dataset. A large number of hidden nodes are used in the network and the number of required hidden nodes is determined through training. We prepared a network with 784 hidden nodes in an auto-associative setting to reconstruct the inputs. The rectified linear unit (ReLU) function is used as the activation function for all the nodes. The sensitivity layer is added after the activation functions and implemented as a collection of 784 individual dense layers, each of which has one input, one output, and one weight. As explained in Section \ref{sec:implementation}, by implementing the sensitivity layer as a special type of dense layer, we can implement and train the network using standard functions in deep learning packages. Here, we used the Keras Python deep learning library for implementation and training. The sensitivity variables are initialized to one. We adopted the $\ell_1$ regularization in the sensitivity layer as a training option and trained the network using the MNIST dataset \cite{lecun1998gradient}.

\begin{figure}[t!]
	\centering		
	\begin{minipage}{0.5\linewidth}		
		\centering
		{\includegraphics[width=\linewidth]{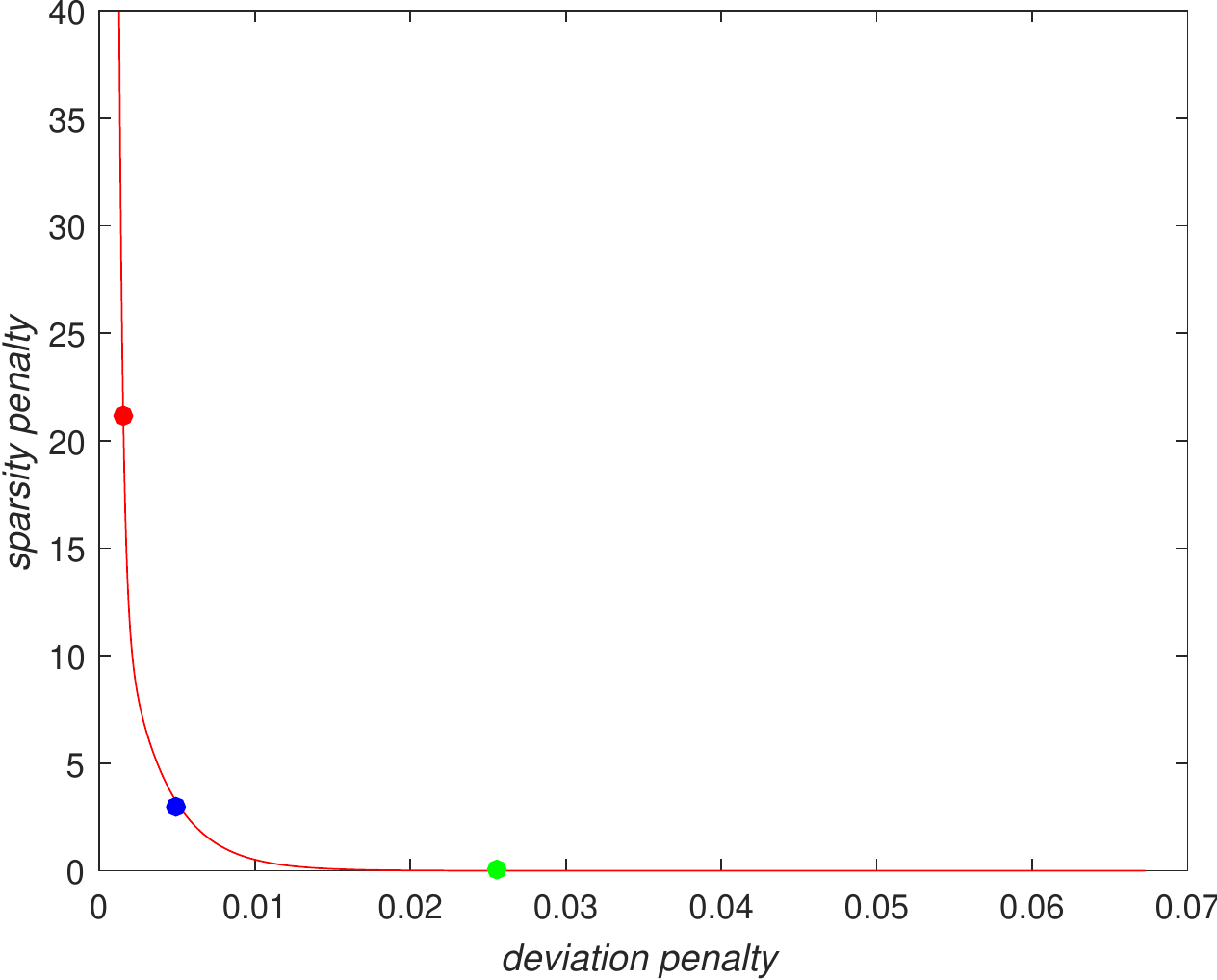}}%
	\end{minipage}%
		
	\caption{L-curve for autoencoder using MNIST dataset, showing $E$ vs. $\sum_{l=1}^L \|\mathbf{s}^l\|_1$ for different values of $\lambda$; red:  $\lambda = 1.0\times 10^{-5}$; blue:  $\lambda = 1.0\times 10^{-3}$; green:  $\lambda = 1.0\times 10^{-1}$. 
	}
	\label{fig:mnistlcurve}
\end{figure}

% L-curve
Optimizing the proposed networks requires the regularization parameter $\lambda$ that weights the deviation and sparsity penalties. We found the appropriate $\lambda$ value using the L-curve. Fig. \ref{fig:mnistlcurve} shows the L-curve for the network using the MNIST dataset, plotted using the deviation penalty $E$ vs. the sparsity $\sum_{l=1}^L \|\mathbf{s}^l\|_1$ at different $\lambda$ values. The linear function is used as the function $\phi$ in \eqref{eq:lcurve}, as the L shape is clearly observed with the choice. For small values of $\lambda$, for example $\lambda = 1.0\times 10^{-5}$, the deviation penalty dominates the cost function of the unconstrained optimization problem in \eqref{eq:optimization2}. Then, the solution to the optimization problem provides only a small deviation penalty but a large sparsity penalty. In contrast, for large values of $\lambda$, for example $\lambda = 1.0\times 10^{-1}$, the sparsity penalty dominates the cost function, providing only a small sparsity penalty but a large deviation penalty. The balance of the two penalties can be achieved using the value from the corner of the L-curve. We used a $\lambda$ value of $1.0\times 10^{-3}$ for the optimization, which corresponds to the corner of the L-curve.

\begin{figure}[t!]
	\centering		
	\begin{minipage}{0.5\linewidth}		
		\centering
		{\includegraphics[width=\linewidth]{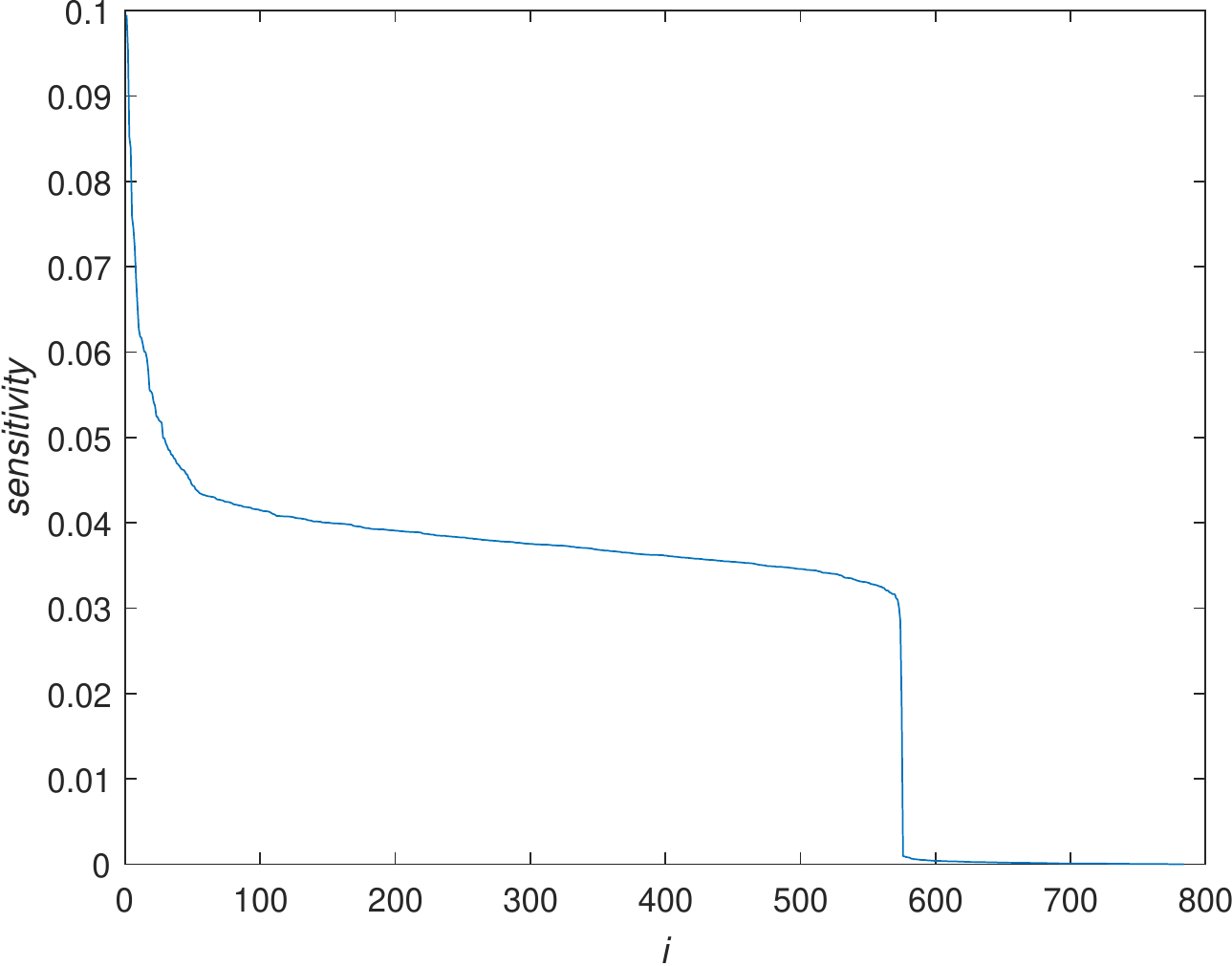}}%
		
		{\footnotesize (a)}
	\end{minipage}%
	\begin{minipage}{0.5\linewidth}		
		\centering
		{\includegraphics[width=\linewidth]{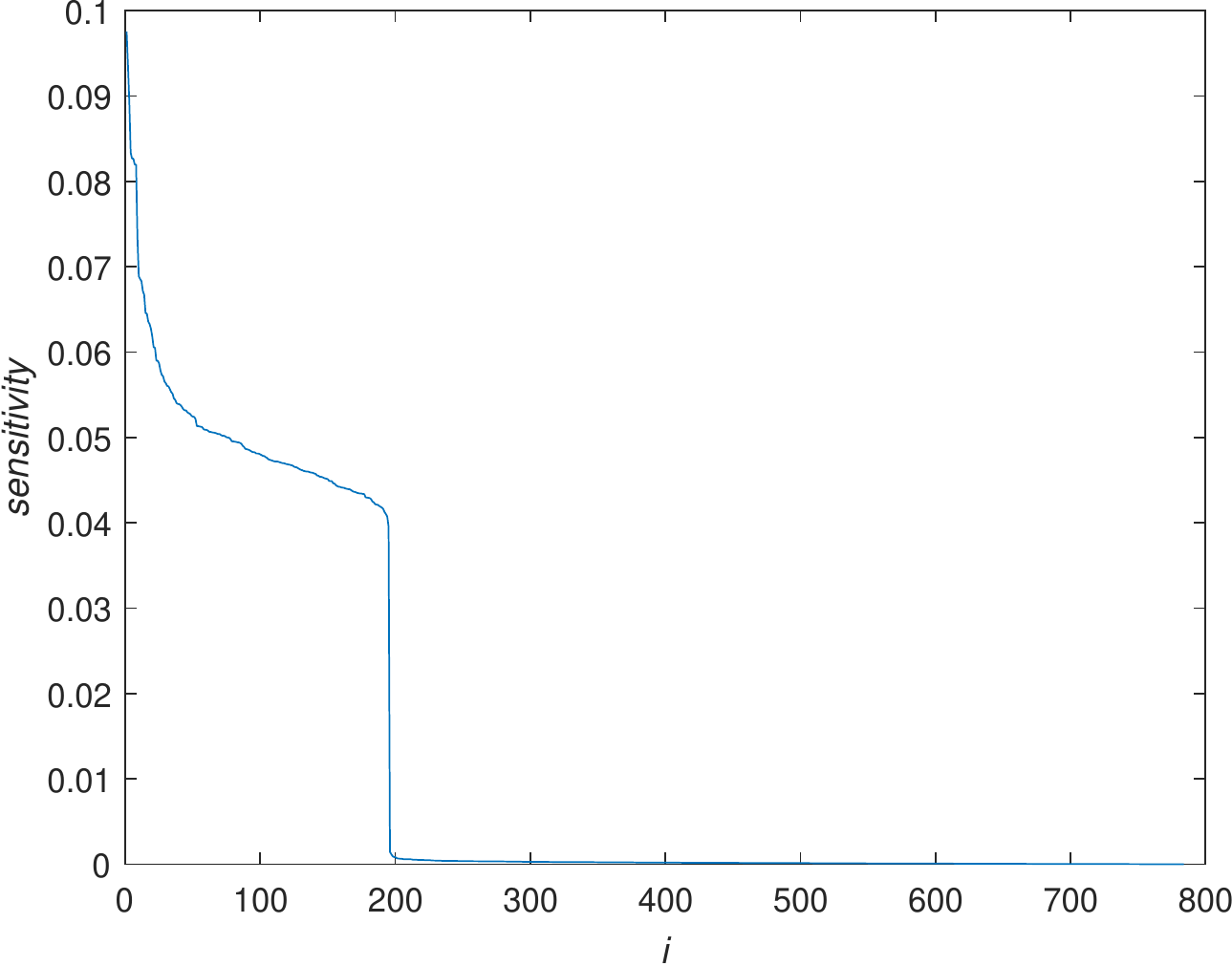}}%
		
		{\footnotesize (b)}
	\end{minipage}%
	
	\begin{minipage}{0.5\linewidth}		
		\centering
		{\includegraphics[width=\linewidth]{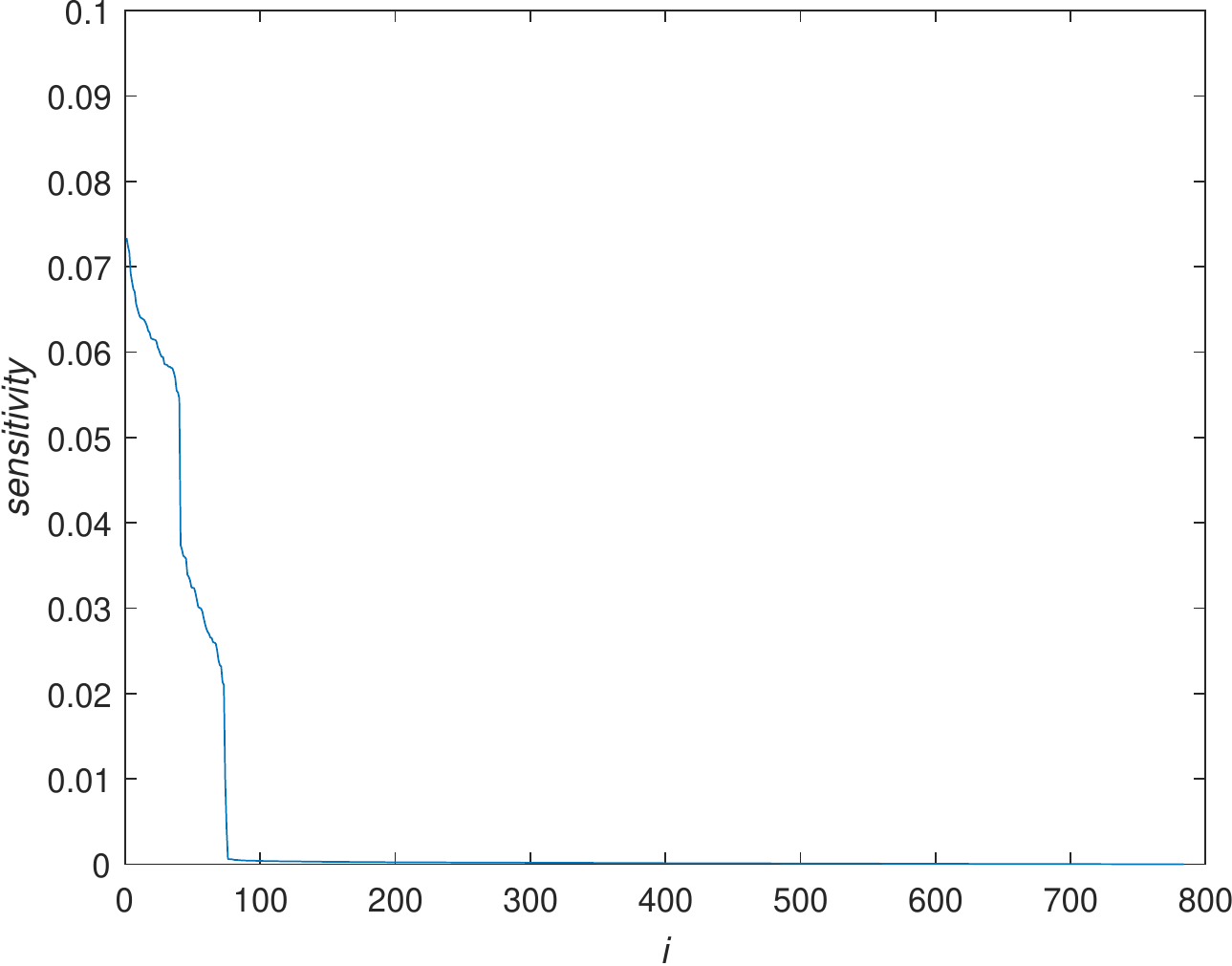}}%
		
		{\footnotesize (c)}
	\end{minipage}%
	\begin{minipage}{0.5\linewidth}		
		\centering
		{\includegraphics[width=\linewidth]{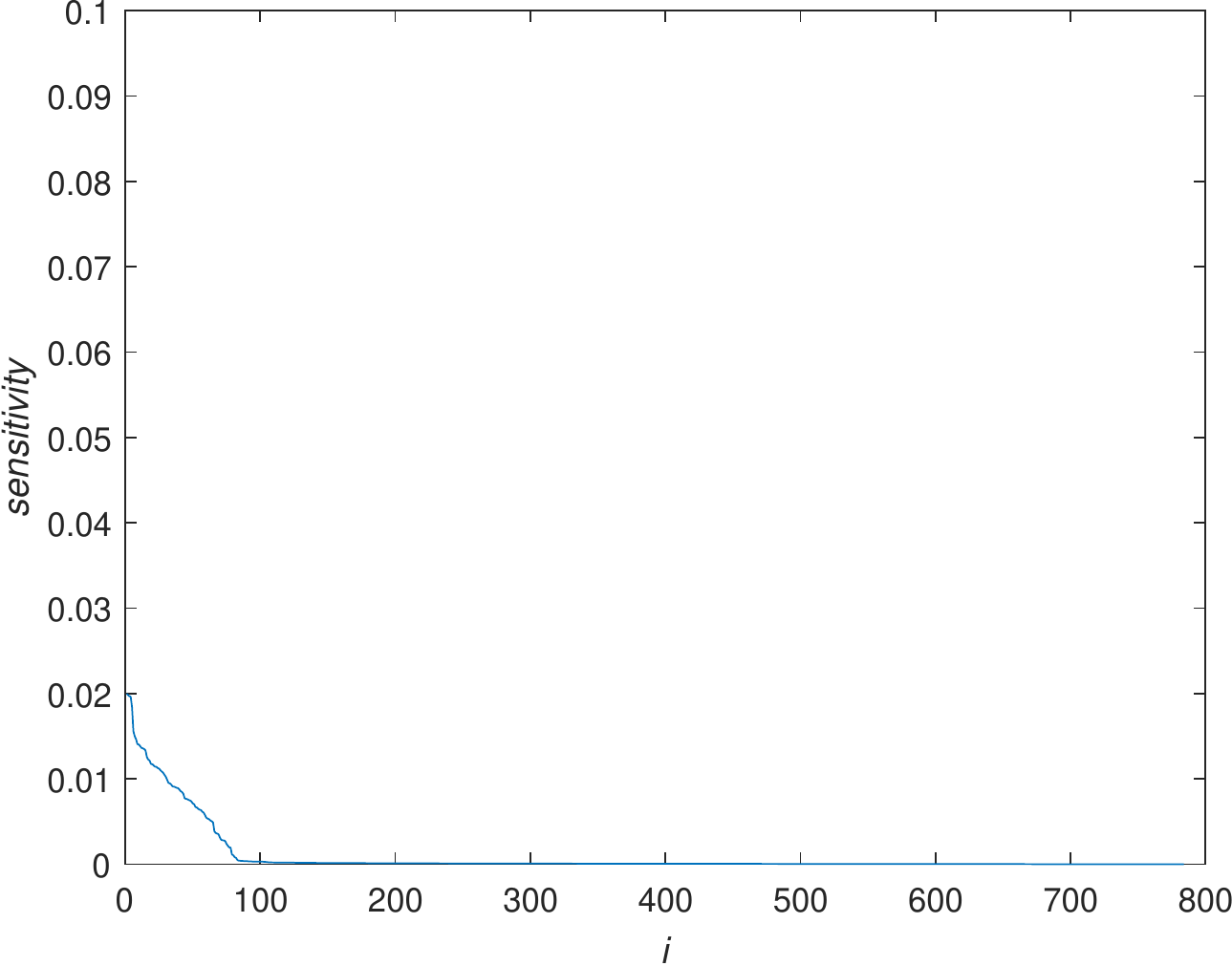}}%
		
		{\footnotesize (d)}
	\end{minipage}%
		
	\caption{Node sensitivity for an autoencoder trained with different $\lambda$ values using the MNIST dataset: 	(a) $\lambda = 1.0\times 10^{-5}$; 	(b) $\lambda = 1.0\times 10^{-4}$; (c) $\lambda = 1.0\times 10^{-3}$; and (d) $\lambda = 1.0\times 10^{-2}$.
	}
	\label{fig:mnistsensitivity}
\end{figure}

% sensitivity
Fig. \ref{fig:mnistsensitivity} shows examples of node sensitivity in networks trained with different values of the regularization parameter $\lambda$. The sensitivity variables, $s_i$, are shown, in the decreasing order, for $\lambda = 1.0\times 10^{-5}, 1.0\times 10^{-4}, 1.0\times 10^{-3}$, and $ 1.0\times 10^{-2}$ in  Fig. \ref{fig:mnistsensitivity} (a), (b), (c), and (d), respectively. As the $\lambda$ value increases, there are more nodes with small sensitivity. When $\lambda=1.0\times 10^{-2}$, many nodes can be removed from the network. However, the trained network fails to provide acceptable performance at this $\lambda$ value. In contrast, a network trained with $\lambda=1.0\times 10^{-3}$ shown in (c) corresponds to the $\lambda$ value at the corner of the L-curve. At this setting, the network provides an acceptable deviation penalty yet has as many zero sensitivity nodes as possible. In effect, the nodes with zero sensitivity are disconnected. The efficient network architecture of the autoencoder with one hidden layer for the MNIST dataset is to have 75 hidden nodes.

% pca
%Fig. \ref{fig:mnistpca} shows the result of PCA on the images in the MNIST dataset. Fig. \ref{fig:mnistpca}(a) shows the eigenvalues in the decreasing order, and Fig. \ref{fig:mnistpca}(b) shows the average mean square error (MSE) values between the inputs and their reconstructions using the $k$ principal components. Fig. \ref{fig:mnistpcanet}(a) shows the node sensitivity in decreasing order, and Fig. \ref{fig:mnistpcanet}(b) shows the average MSE values between the inputs and their reconstructions using the $k$ most sensitive nodes. A network with the sensitivity layer operates similarly to PCA. The MSE values do not change by removing the nodes with zero sensitivity. Moreover, the MSE values increase gradually when nodes are removed in the order of sensitivity. If we needed to remove more nodes than just the ones with zero sensitivity, we could use the sensitivity variables to determine which nodes to remove. 

\begin{figure}[t!]
	\centering		
	\begin{minipage}{0.5\linewidth}		
		\centering
		{\includegraphics[clip, width=\linewidth]{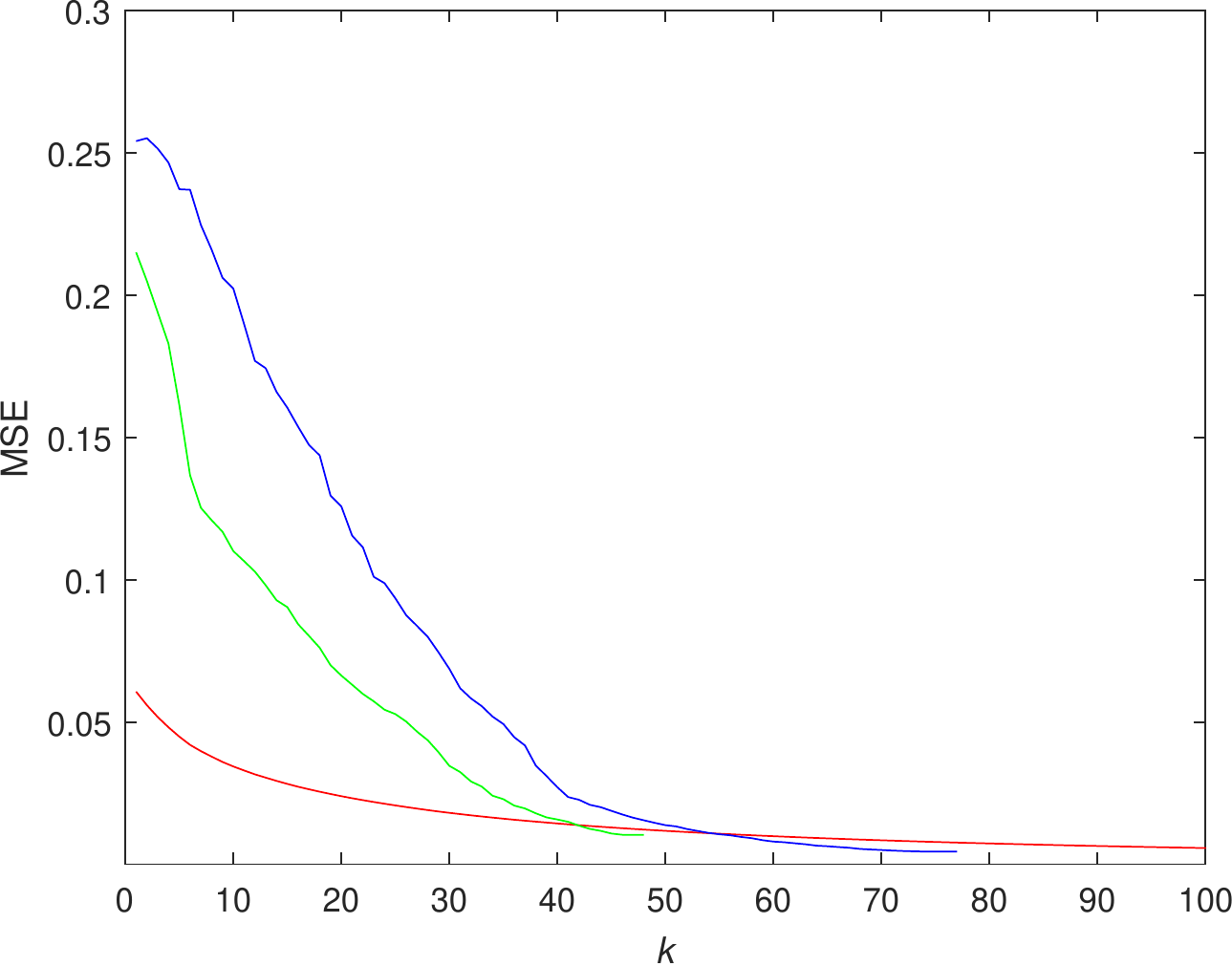}}%
		
	\end{minipage}%
		
	\caption{
	PCA of images in MNIST dataset, average MSE of reconstructed images:
	red: when the first $k$ principal components are used;
	blue: when the first $k$ nodes in the network with heterogenous sensitivity are used ($\lambda=1.0\times 10^{-3}$);
	green: when the first $k$ nodes in the network with heterogenous sensitivity are used ($\lambda=6.0\times 10^{-3}$).
	}
	
	\label{fig:mnistpca}
\end{figure}

Fig. \ref{fig:mnistpca} shows the result of PCA on the images in the MNIST dataset. The average mean square error (MSE) values between the inputs and their reconstructions using the $k$ principal components are shown in red line. The average MSE values between the inputs and their reconstructions using the $k$ most sensitive nodes are shown in blue line for the network trained with $\lambda=1.0\times 10^{-3}$. This network has 75 nodes. When all the 75 nodes are used to reconstruct the images, the MSE is lower than one would achieve by reconstructing the images using the 75 principal components. However, when fewer than 75 nodes are used, the MSE values degrade faster than the PCA results. The average MSE values between the inputs and their reconstructions for the network trained with $\lambda=6.0\times 10^{-3}$ are shown in green line. This network has 46 nodes. When all the 46 nodes are used to reconstruct the images, the MSE is lower than one would achieve by reconstructing the images using the 46 principal components. This observation suggests that a network with a fewer number of nodes should be designed by training a network using a higher regularization parameter value rather than removing nodes from a trained network.

\subsection{Regularzation Parameter Selection}
\label{sec:regularization}

In Section \ref{sec:lcurve}, an algorithm to determine the regularization parameter simultaneously with the training of a network is proposed. In this section, we train a network using Algorithm \ref{alg:training}, and compare the networks performance and complexity to a network trained using the regularization parameter at the corner of the L-curve.

\begin{figure}[t!]
	\centering		
	\begin{minipage}{0.5\linewidth}		
		\centering
		{\includegraphics[clip, width=\linewidth]{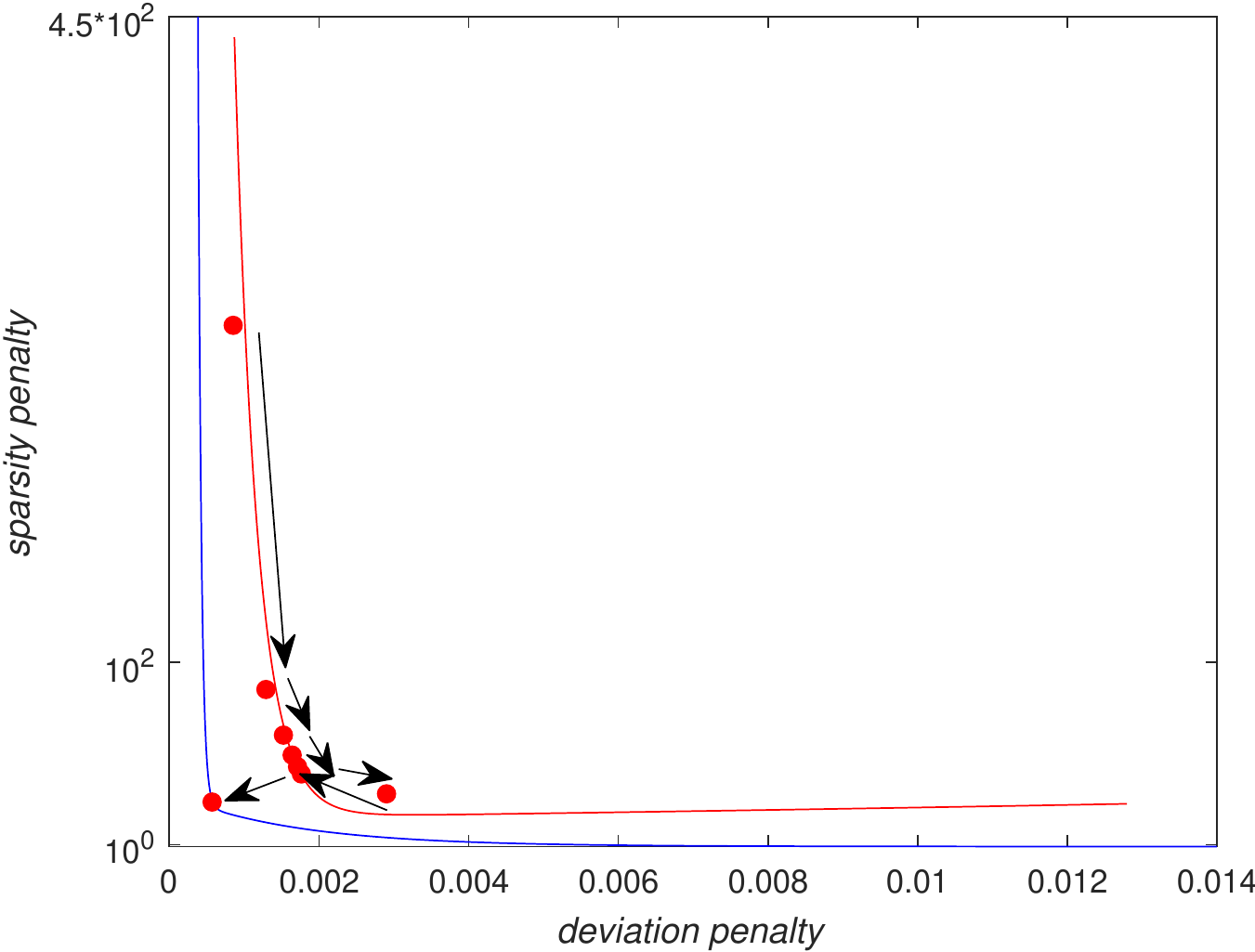}}%
		
	\end{minipage}%
		
	\caption{
		Simultaneous regularization parameter selection and network training:
		red: L-curve constructed by Algorithm 1;
		blue: L-curve constructed by training the same network multiple times at different regularization parameters.
	}
	\label{fig:simultaneous}
\end{figure}

The same network used in the previous section, the autoencoder with a hidden layer, is trained using the Algorithm \ref{alg:training}. Fig. \ref{fig:simultaneous} shows the deviation and sparsity penalties at different values of the regularization parameter used during the training by the Algorithm \ref{alg:training}. The algorithm increases the regularization parameters until the deviation penalty during the training increases considerable. Then, the algorithm trains the network using the selected regularization parameter until the termination condition is met. The L-curve constructed by training the same network multiple times at different regularization parameters is also shown. It can be seen that the deviation and sparsity penalty of the trained network is close to the corner point of the L-curve.

The regularization parameter simultaneously selected during the training by the Algorithm \ref{alg:training} is $5.0\times 10^{-4}$, while the regularization parameter that corresponds to the corner of the L-curve is $1.0\times 10^{-3}$. The numbers of hidden nodes are 89 and 75 and the training loss of the network is 0.0018 and 0.0009 for the network trained by the Algorithm \ref{alg:training} and the one trained with the L-curve corner value. The results obtained by determining the regularization parameter via the Algorithm \ref{alg:training} were the same as one would get by hand-selecting the parameter corresponding to the corner of the L-curve.

We used the Algorithm \ref{alg:training} using four epochs for each updated values of $\lambda$, and the algorithm updated seven $\lambda$ values. After the $\lambda$ values is found, the network is trained for 268 epochs before the termination of the training. For comparison, the network with a fixed values of $\lambda$ is trained for 279 epochs. The simultaneous training and selection of $\lambda$ by the Algorithm \ref{alg:training} requires 6.1\% more epochs than the training for a single values of $\lambda$.

\subsection{Deep CNN with CIFAR-10 Dataset}
\label{ex:cnncifar}

% setup
In this experiment, we consider a deep CNN with heterogeneous sensitivity for object classification using the CIFAR-10 dataset \cite{krizhevsky2009learning}. We prepared a network with four convolutional layers and two dense layers, adding sensitivity layers to all the convolutional layers and the first dense layers. The second dense layer is designed to perform object classification. The sensitivity layers for the convolutional layers are implemented as a collection of convolutional layers, each of which has one input node and one output node with a one-by-one filter. The sensitivity layers for the dense layers are implemented as a collection of dense layers, each of which has one input node, one output node, and one weight. We applied $\ell_1$ regularization in the sensitivity layers for training. For comparison, we trained a baseline CNN with the same number of nodes and layers and with the ReLU activation function using the same training set. The batch normalization is applied after the convolutional layers and the dropout is applied after the first dense layer on both the asymmetric and the baseline symmetric networks.

\begin{figure}[t!]
	\centering		
	\begin{minipage}{0.5\linewidth}		
		\centering
		{\includegraphics[width=\linewidth]{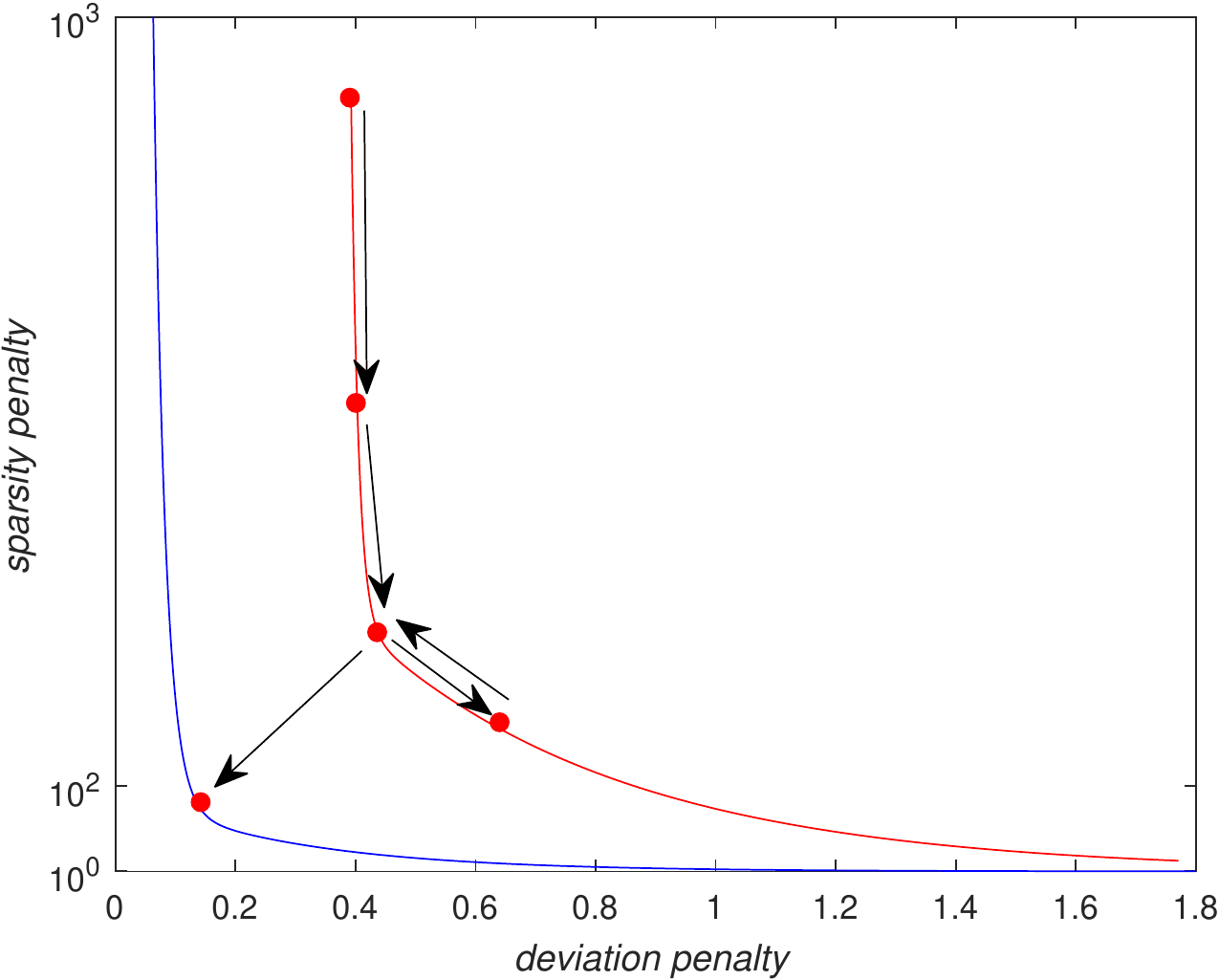}}%
	\end{minipage}%
		
	\caption{L-curve for CNN with CIFAR-10 data, showing $E$ vs. $\sum_{l=1}^L \|\mathbf{s}^l\|_1$ at different values of $\lambda$;
			red: L-curve constructed by Algorithm 1;
			blue: L-curve constructed by training the same network multiple times at different regularization parameters.	
	}
	\label{fig:cnnlcurve}
\end{figure}

\begin{figure}[t!]
	\centering		
	\begin{minipage}{0.5\linewidth}		
		\centering
		{\includegraphics[width=\linewidth]{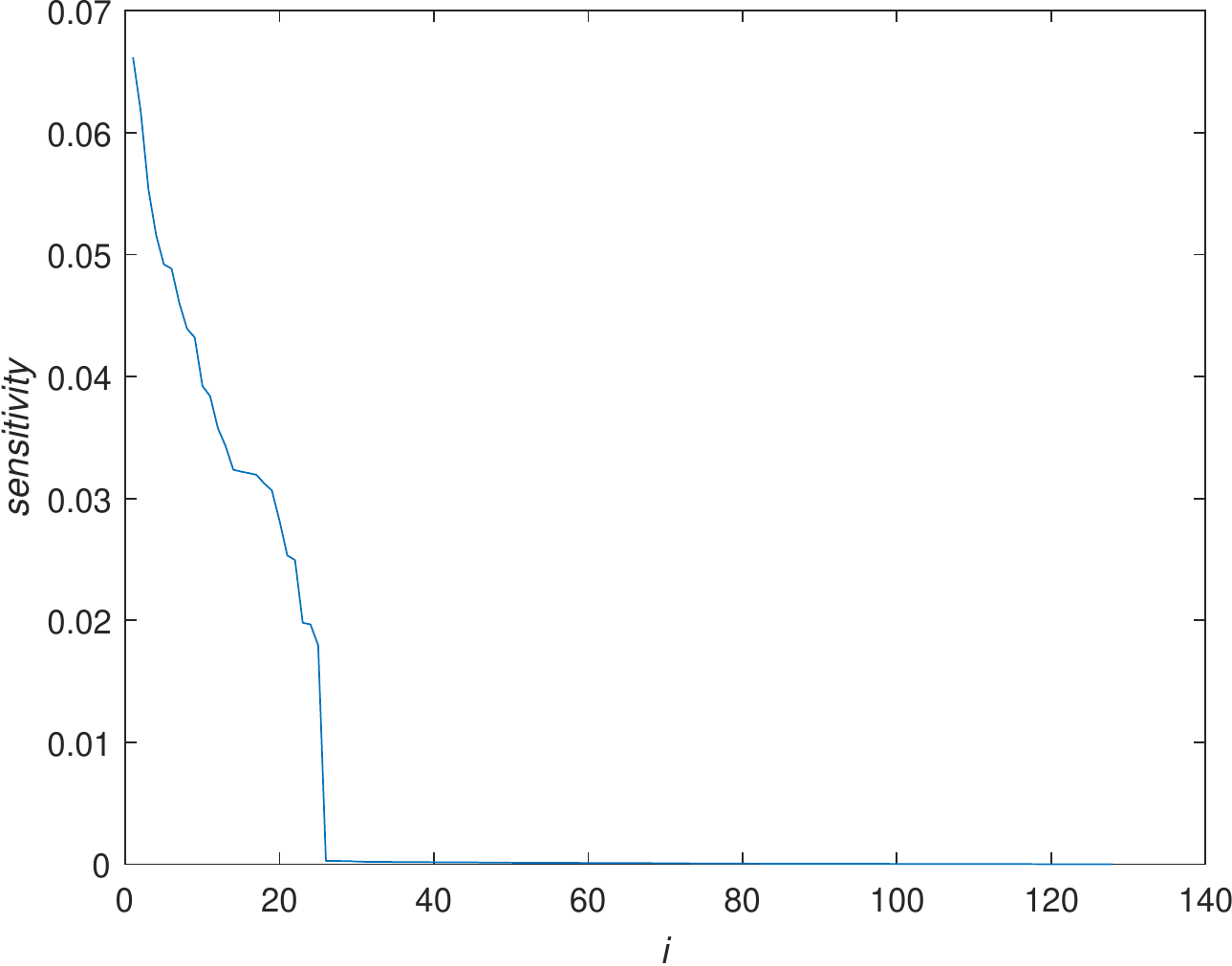}}%
		
		{\footnotesize (a)}
	\end{minipage}%
	\begin{minipage}{0.5\linewidth}		
		\centering
		{\includegraphics[width=\linewidth]{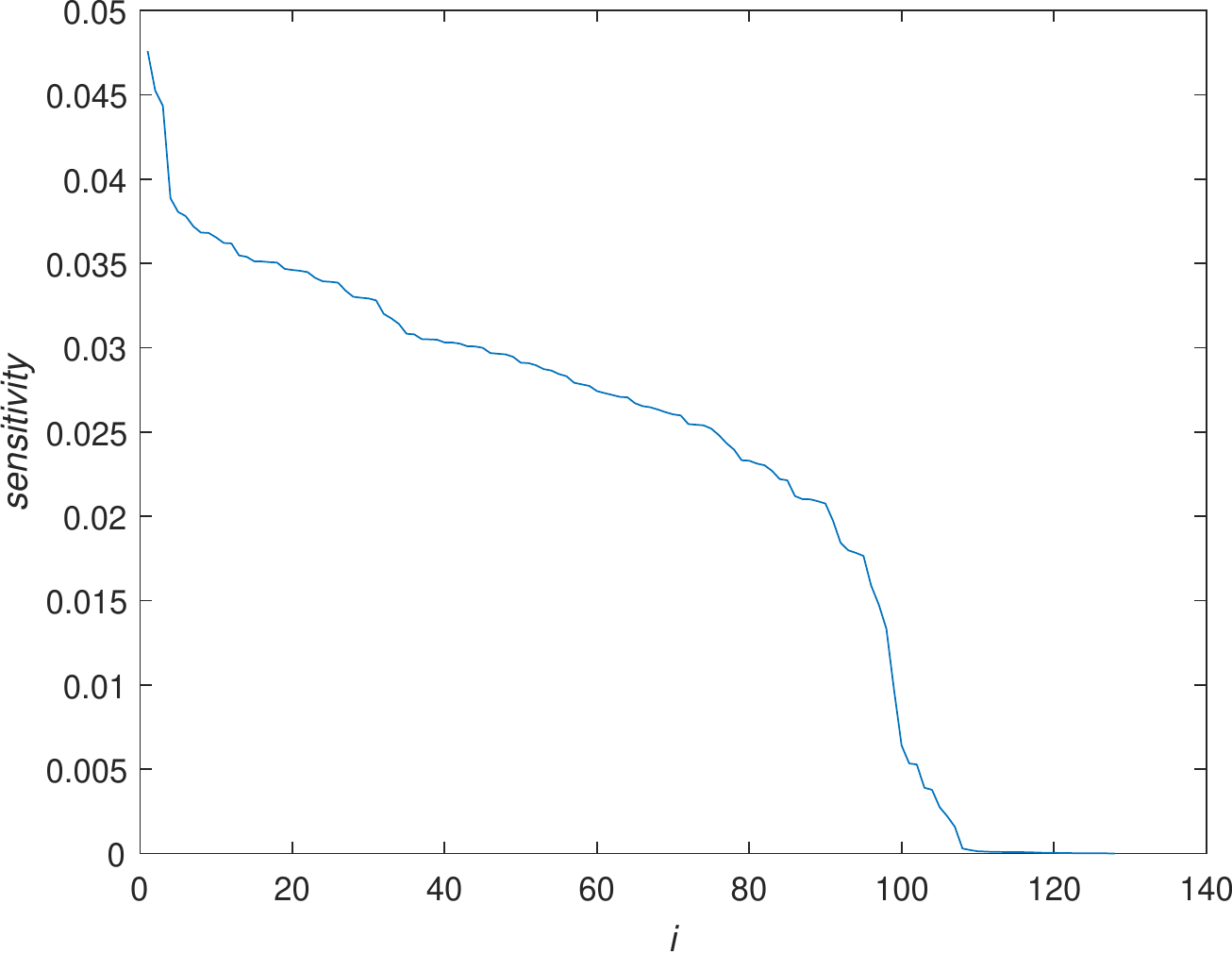}}%
		
		{\footnotesize (b)}
	\end{minipage}%
	
	\begin{minipage}{0.5\linewidth}		
		\centering
		{\includegraphics[width=\linewidth]{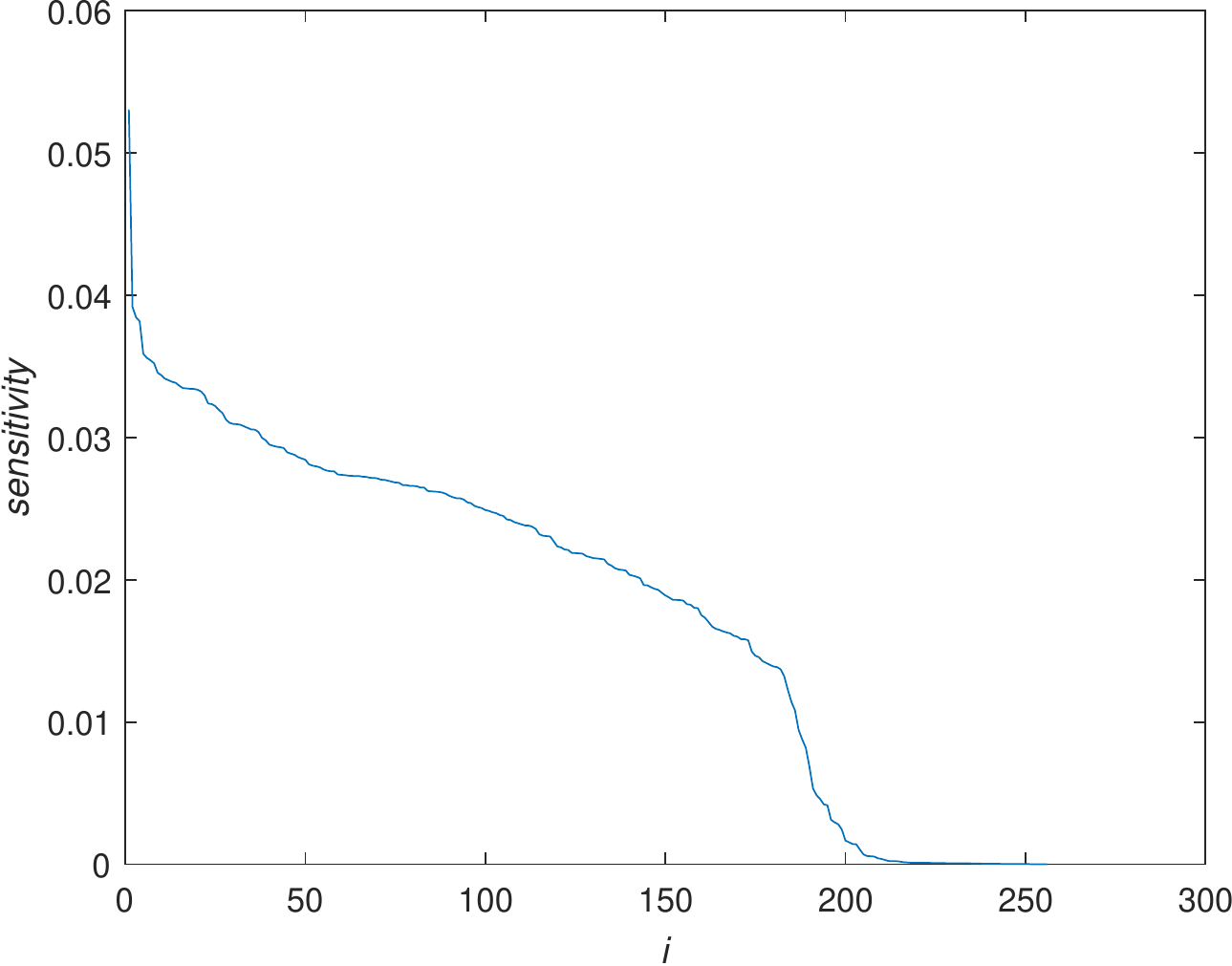}}%
		
		{\footnotesize (c)}
	\end{minipage}%
	\begin{minipage}{0.5\linewidth}		
		\centering
		{\includegraphics[width=\linewidth]{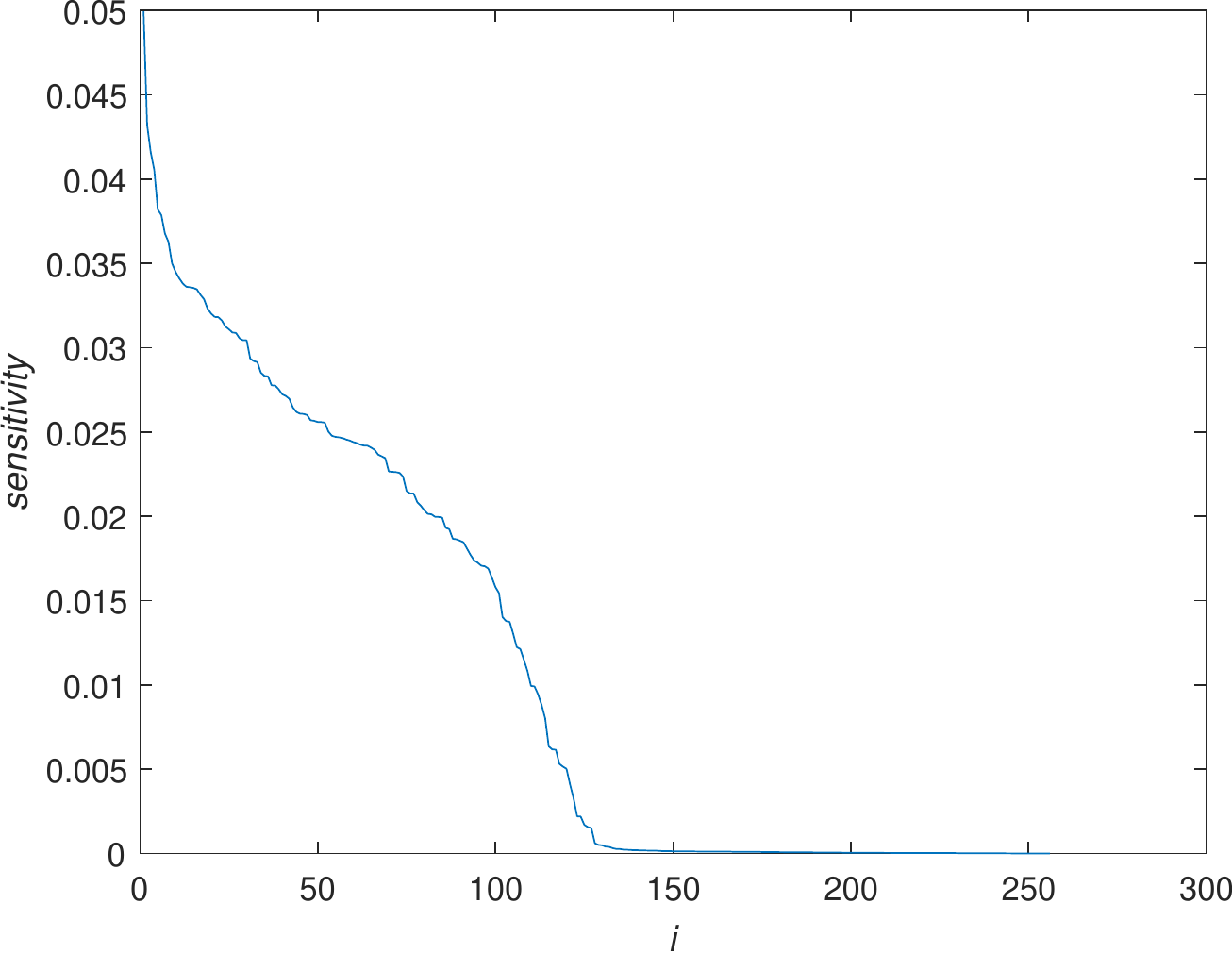}}%
		
		{\footnotesize (d)}
	\end{minipage}%

	\begin{minipage}{0.5\linewidth}		
		\centering
		{\includegraphics[width=\linewidth]{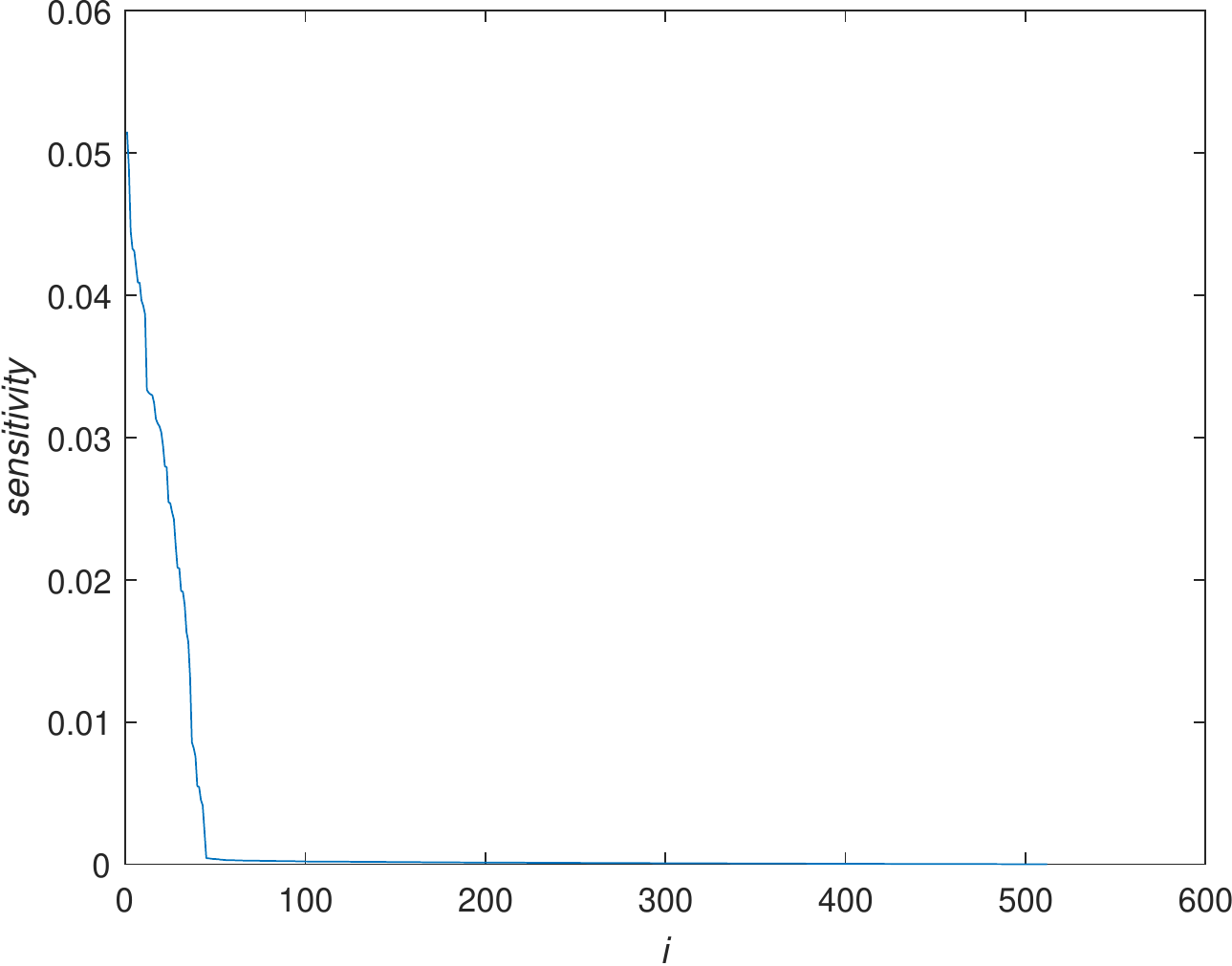}}%
		
		{\footnotesize (e)}
	\end{minipage}%
	\begin{minipage}{0.5\linewidth}		
		\centering
		{\includegraphics[width=\linewidth]{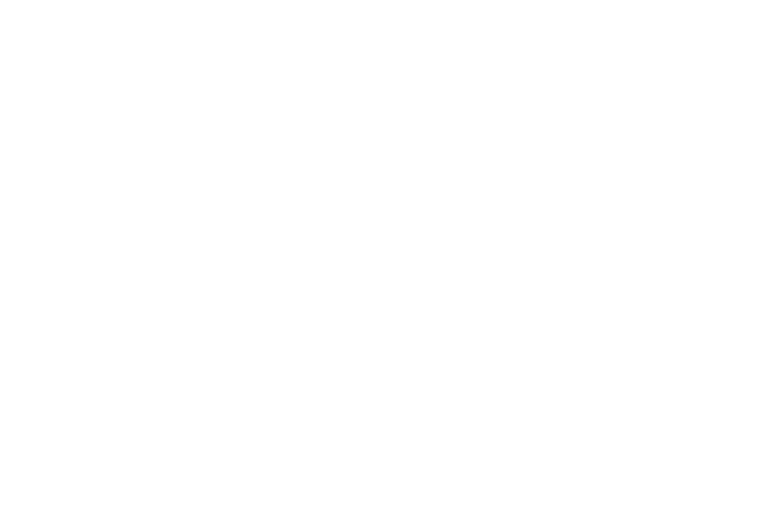}}%

	\end{minipage}%

	\caption{Sensitivity variables for the proposed CNN with CIFAR-10 data: 	(a) 1st conv layer; (b) 2nd conv layer; (c) 3rd conv layer; 	(d) 4th conv layer; and 	(e) 1st dense layer.
	}
	\label{fig:cnnsensitivity}
\end{figure}

% results
Fig. \ref{fig:cnnlcurve} shows the L-curve for the CNN. The regularization parameter simultaneously selected during the training by the Algorithm \ref{alg:training} is $0.7\times 10^{-3}$, while the regularization parameter that corresponds to the corner of the L-curve is $0.8\times 10^{-3}$. Fig. \ref{fig:cnnsensitivity} shows the sensitivity of the nodes, in decreasing order, in each layer of the network trained with the find the regularization parameter. The sensitivity variables $s^l_i$ are sparse in all the layers and have many zero elements. Only the nodes with non-zero sensitivity need to be included in the efficient architecture. We used a thresholding approach to remove the nodes with sensitivity values numerically close to zero to find the efficient architecture.

% complexity
We used the Algorithm \ref{alg:training} for the simultaneous regularization parameter selection using four epochs for each updated values of $\lambda$, and the algorithm updated five $\lambda$ values. After the $\lambda$ values is found, the network is trained for 215 epochs before the termination of the training. For comparison, the network with a fixed values of $\lambda$ is trained for 221 epochs. The simultaneous training and selection of $\lambda$ by the Algorithm \ref{alg:training} requires 6.3\% more epochs than the training for a single values of $\lambda$.

% results
Table \ref{tab:cnncifar} summarizes the complexity and performance of the proposed efficient CNN compared to the baseline CNN. The proposed network includes only $39.30\%$ of the nodes and $37.71\%$ of the weights, and requires $33.21\%$ of the FLOPs compared to the baseline CNN. The accuracy of the efficient and baseline CNN are $83.20\%$ and $83.12\%$, respectively. By using nodes with heterogeneous sensitivity, the CNN learns to classify objects with the same accuracy as the baseline but uses only an optimal number of nodes. We included the layer-wise number of nodes, weights, and FLOPs data in the supporting materials.

\begin{table}[t!]
	\centering
	\caption{Performances of Proposed Efficient CNN with CIFAR-10 data}
	\label{tab:cnncifar}
\begin{tabular}{l|rrr}								
\hline								
	&	baseline	&	proposed	&	ratio	\\	\hline
\# of nodes	&	1290	&	507	&	39.30\%	\\	
\# of weights	&	3407498	&	1284942	&	37.71\%	\\	
\# of FLOPs	&	1.93E+08	&	6.43E+07	&	33.21\%	\\	\hline
accuracy	&	83.20\%		&	83.12\%	&	99.90\%	\\	\hline
\end{tabular}																
\end{table}

\subsection{VGG and ResNet with CK+ Dataset}
\label{ex:vggresnetck}

% set up
A trained network can be transferred to form a basis for the design of a network intended for another task. In this section, we added the sensitivity layers to transferred networks to find efficient architectures for a different task. We tested transferred VGG \cite{simonyan2014very} and ResNet \cite{he2016deep} for facial expression recognition using the CK+ dataset  \cite{lucey2010extended} to classify input facial images into seven emotions:
\begin{equation}
	\{\hbox{anger, contempt, disgust, fear, surprise, happiness, sadness}\}.
	\nonumber
\end{equation} 
We selected 325 sequences of 118 subjects that are classified as displaying one of the seven emotions. The so-called ``apex frames'' that occur at the peak of the expression were collected as labeled facial images. The network was trained and tested using the ten-fold cross validation protocol.

% setup
The VGG-16 and ResNet-56 networks with heterogeneous sensitivity were prepared by adding the sensitivity layers after all the convolutional and dense layers. The batch normalization and the dropout are used for the networks. The networks were trained using the CK+ facial recognition dataset. We determined the regularization parameter $\lambda$ using the L-curve as described earlier. After the training, we included only the nodes with non-zero sensitivity in the efficient architecture. For comparison, the baseline VGG-16 and ResNet-56 are trained using the same training set.

% result

Table \ref{tab:vggck} summarizes the complexity and performance of the proposed efficient VGG-16 compared to the baseline VGG-16. The efficient network includes only $12.41\%$ of the nodes and $1.11\%$ of the weights, and requires $3.07\%$ of the FLOPs compared to the baseline VGG-16. The accuracy of the efficient and baseline VGG-16 are $97.89\%$ and $97.85\%$, respectively. Table \ref{tab:resnetck} summarizes the complexity and performance of the efficient ResNet-56 compared to the baseline ResNet-56. The proposed efficient network includes only $12.64\%$ of the nodes and $1.94\%$ of the weights, and requires $17.53\%$ of the FLOPs compared to the baseline ResNet-56. The accuracy of the efficient and baseline ResNet are $96.77\%$ and $96.03\%$, respectively. By using nodes with heterogeneous sensitivity, the transferred VGG and ResNet learns to classify facial expressions with the same accuracy as the baseline but uses only a small number of nodes. We included the node-wise sensitivity variables and the number of nodes, weights, and FLOPs data in the supporting materials.

\begin{table}[t!]
	\centering
	\caption{Complexity and Performance of Proposed Efficient VGG-16 with the CK+ Dataset}
	\label{tab:vggck}
\begin{tabular}{l|rrr}								
\hline								
	&	baseline	&	proposed	&	ratio	\\	\hline
\# of nodes	&	4743	&	589	&	12.41\%	\\	
\# of weights	&	15767367	&	174327	&	1.11\%	\\	
\# of FLOPs	&	1.25E+09	&	3.85E+08	&	3.07\%	\\	\hline
accuracy	&	97.89\%	&	97.85\%	&	100.00\%	\\	\hline
\end{tabular}								
\end{table}

\begin{table}[t!]
	\centering
	\caption{Performances of Proposed Efficient ResNet with the CK+ Dataset}
	\label{tab:resnetck}
\begin{tabular}{l|rrr}										
\hline										
	&	baseline	&	proposed	&	ratio	\\	\hline		
\# of nodes	&	27207	&	3441	&	12.64\%	\\			
\# of weights	&	31872135	&	619369	&	1.94\%	\\			
\# of FLOPs	&	1.33E+09	&	2.86E+08	&	17.53\%	\\	\hline		
accuracy	&	96.77\%	&	96.03\%	&	99.2\%	\\	\hline		
\end{tabular}										
\end{table}

\subsection{YOLO with VOC Dataset}
\label{ex:yolovoc}

% setup
We prepared a YOLO network with heterogeneous sensitivity by adding sensitivity layers after all the convolutional layers. We simplified the object detection task by considering only four object classes: car, motorbike, pedestrian, and people. The network was trained using the VOC dataset. The regularization parameter $\lambda$ was found using the L-curve as described previously. 

% results
Table \ref{tab:yolovoc} summarizes the complexity and performance of the proposed efficient YOLO compared to the baseline YOLO. The efficient network includes only $56.69\%$ of the nodes and $26.18\%$ of the weights, and requires $43.12\%$ of the FLOPs compared to the baseline YOLO. The accuracy of the efficient and baseline YOLO in terms of the mean average precision (mAP) are 70.9 and 69.2, respectively. By using nodes with heterogeneous sensitivity, the YOLO learns to detect the specified objects with the same accuracy as the baseline but uses only a small number of nodes. We included the node-wise sensitivity variables and the number of nodes, weights, and FLOPs data in the supporting materials.

\begin{table}[t!]
	\centering
	\caption{Performance of Proposed Efficient YOLO with the VOC Dataset}
	\label{tab:yolovoc}
\begin{tabular}{l|rrr}								
\hline								
	&	baseline	&	proposed	&	ratio	\\	\hline
\# of nodes	&	10381	&	5885	&	56.69\%	\\	
\# of weights	&	50594061	&	13244927	&	26.18\%	\\	
\# of FLOPs	&	1.54E+10	&	6.66E+09	&	43.12\%	\\	\hline
mAP	&	70.9	&	69.2	&	97.60\%	\\	\hline
\end{tabular}																								
\end{table}

\subsection{LeNet with MNIST Dataset}
\label{ex:lenetmnist}

% comparisons
We compared the complexity and performance of an proposed efficient network found by the proposed method to the pruning results reported in \cite{han2015learning, zhou2016less, yang2015deep, lebedev2016fast, srinivas2015data, jang2018deep}. The LeNet-5 \cite{krizhevsky2012imagenet} network with heterogeneous sensitivity was prepared and trained using the MNIST dataset. The ratios of the weights remaining after the optimization and the classification errors are reported in Table \ref{tab:compMNIST}. The efficient network designed by the proposed method provides the highest performance with the least computational complexity.

\begin{table}[t!]
	\centering
	\caption{Comparison to Pruning Methods Reported with LeNet on MNIST Dataset}
	\label{tab:compMNIST}
\begin{tabular}{clrrr}										
\hline										
	&		&	\# of FLOPs	&	\# of weights	&	error	\\	\hline
\cite{han2015learning}	&	baseline	&		&		&	0.80\%	\\	
	&	pruned	&	16.0\%	&	8.24\%	&	0.77\%	\\	\hline
\cite{zhou2016less}	&	baseline	&		&		&	0.73\%	\\	
	&	pruned	&	N/A 	&	10.25\%	&	0.76\%	\\	\hline
\cite{yang2015deep}	&	baseline	&		&		&	0.87\%	\\	
	&	pruned	&	12.1\%	&	9.01\%	&	0.71\%	\\	\hline
\cite{lebedev2016fast}	&	baseline	&		&		&	N/A	\\	
	&	pruned	&	N/A 	&	8.33\%	&	1.70\%	\\	\hline
\cite{srinivas2015data}	&	baseline	&		&		&	0.94\%	\\	
	&	pruned	&	16.5\%	&	16.00\%	&	1.65\%	\\	\hline
\cite{jang2018deep}	&	baseline	&		&		&	0.81\%	\\	
	&	pruned	&	78.6\%	&	6.73\%	&	0.71\%	\\	\hline
proposed	&	baseline	&		&		&	0.80\%	\\	
	&	proposed	&	16.0\%	&	6.40\%	&	0.69\%	\\	\hline
\end{tabular}														
\end{table}

\subsection{VGG and ResNet with CIFAR-10 Dataset}
\label{ex:vggresnetcifar}

% comparisons
We compared the complexity and performance of the proposed efficient networks with the pruning results reported in \cite{li2016pruning, wen2016learning, jang2018deep, ayinde2018building}. The VGG-16, ResNet-56, ResNet-20 networks with heterogeneous sensitivity were prepared using the CIFAR-10 dataset. The batch normalization and the dropout are used for the networks. The ratios of the weights remaining after the optimization and the classification errors are reported in Table \ref{tab:cifarcomp}. The efficient networks designed by the proposed method provide the same or higher performance but with less computational complexity.

\begin{table}[t!]
	\centering
	\caption{Comparison to Pruning Methods Reported with VGG and ResNet on the CIFAR-10 Dataset}
	\label{tab:cifarcomp}
	
\begin{tabular}{cllrrr}												
\hline												
	&		&		&	\# of FLOPs	&	\# of weights	&	error	\\	\hline
\cite{li2016pruning}	&	VGG-16	&	baseline	&		&		&	6.75\%	\\	
	&		&	pruned	&	65.6\%	&	36.0\%	&	6.60\%	\\	\hline
\cite{jang2018deep}	&	VGG-16	&	baseline	&		&		&	6.67\%	\\	
	&		&	pruned	&	67.8\%	&	29.6\%	&	6.17\%	\\	\hline
\cite{ayinde2018building}	&	VGG-16	&	baseline	&		&		&		\\	
	&		&	pruned	&	59.4\%	&		&	6.33\%	\\	\hline
proposed	&	VGG-16	&	baseline	&		&		&	6.67\%	\\	
	&		&	proposed	&	52.9\%	&	16.1\%	&	6.52\%	\\	\hline
\cite{li2016pruning}	&	ResNet-56	&	baseline	&		&		&	6.96\%	\\	
	&		&	pruned	&	72.6\%	&	86.3\%	&	6.94\%	\\	\hline
\cite{he2017channel}	&	ResNet-56	&	baseline	&		&		&	7.20\%	\\	
	&		&	pruned	&	50.0\%	&	N/A	&	8.20\%	\\	\hline
\cite{jang2018deep}	&	ResNet-56	&	baseline	&		&		&	6.82\%	\\	
	&		&	pruned	&	18.4\%	&	26.2\%	&	6.75\%	\\	\hline
\cite{ayinde2018building}	&	ResNet-56	&	baseline	&		&		&		\\	
	&		&	pruned	&	72.6\%	&	76.4\%	&	6.88\%	\\	\hline
proposed	&	ResNet-56	&	baseline	&		&		&	6.82\%	\\	
	&		&	proposed	&	45.4\%	&	40.2\%	&	6.70\%	\\	\hline
\cite{wen2016learning}	&	ResNet-20	&	baseline	&		&		&	8.82\%	\\	
	&		&	pruned	&	N/A 	&	N/A	&	7.51\%	\\	\hline
Proposed	&	ResNet-20	&	baseline	&		&		&	8.21\%	\\	
	&		&	proposed	&	66.2\%	&	72.3\%	&	6.76\%	\\	\hline
\end{tabular}																																				
\end{table}

\subsection{ResNet with the ImageNet Dataset}
\label{ex:resnetimagenet}

% comparisons
The ResNet-50 network with heterogeneous sensitivity were prepared using the ImageNet dataset. The batch normalization and the dropout are used for the network. With the ImageNet dataset, the network converges very slowly with inconsistent improvement of the penalty function during the training. The regularization parameter is selected empirically and trained with the selected parameter. The ratios of the weights remaining after the optimization and the classification errors are reported in Table \ref{tab:compIMAGENET} where comparisons to the pruning results reported in \cite{luo2017thinet, luo2017entropy, zhuang2018discrimination, xu2018hybrid, he2017channel} are also given. The efficient networks designed by the proposed method provide the same or higher performance but with less computational complexity.

\begin{table}[t!]
	\centering
	\caption{Comparison to Pruning Methods Reported with ResNet on The ImageNet Dataset}
	\label{tab:compIMAGENET}

\begin{tabular}{clrrr}										
\hline										
	&		&	\# of FLOPs	&	\# of weights	&	accuracy	\\	\hline
\cite{luo2017thinet}	&	baseline	&		&		&	72.88\%	\\	
	&	pruned (ThiNet-70)	&	63.2\%	&	63.3\%	&	72.04\%	\\	
	&	pruned (ThiNet-50)  &	44.2\%	&	48.4\%	&	71.01\%	\\	
	&	pruned (ThiNet-30)	&	28.4\%	&	33.9\%	&	68.42\%	\\	\hline
\cite{luo2017entropy}	&	baseline	&		&		&	72.88\%	\\	
	&	pruned (Pruned-90)	&	92.7\%	&	93.5\%	&	73.56\%	\\	
	&	pruned (Pruned-75)	&	82.6\%	&	84.9\%	&	72.89\%	\\	
	&	pruned (Pruned-50)	&	65.2\%	&	68.0\%	&	70.84\%	\\	\hline
\cite{zhuang2018discrimination}	&	baseline	&		&		&	76.01\%	\\	
	&	pruned (DCP)	&	44.4\%	&	48.5\%	&	73.20\%	\\	
	&	pruned (WM+)	&	44.4\%	&	48.5\%	&	72.89\%	\\	
	&	pruned (WM)		&	44.4\%	&	48.5\%	&	70.84\%	\\	\hline
\cite{xu2018hybrid}	&	baseline	&		&		&	76.01\%	\\	
	&	pruned	&	N/A 	&	67.3\%	&	74.87\%	\\	\hline
\cite{he2017channel}	&	baseline	&		&		&	75.30\%	\\	
	&	pruned	&	N/A 	&	64.0\%	&	72.30\%	\\	\hline
Proposed	&	baseline	&		&		&	75.06\%	\\	
	&	proposed	&	68.6\%	&	56.3\%	&	75.03\%	\\	\hline
\end{tabular}										

\end{table}
\section{Conclusion}
\label{sec:conclusion}

In this study, we trained networks consisting of nodes with heterogeneous sensitivity to perform a given task using only a small number of sensitive nodes. The training is formulated as a constrained optimization problem whose parameter is found simultaneously during the training based on the L-curve. By introducing sensitivity layers that assign sensitivity variables to nodes, we were able to implement and train a network without using a complicated optimization tool.  The networks trained in this manner possess a small and computationally efficient network architecture and simultaneously meet the performance criteria. In our experiments, the efficient networks designed by the proposed method provide the same or higher performance but with far less computational complexity. The proposed method can be used to determine the efficient network architectures of deep networks.

\bibliographystyle{IEEEtran}
\bibliography{IEEEabrv,./a.bib}

\clearpage

\begin{IEEEbiographynophoto}{Hyunjoong Cho} received a B. S. degree in 2015 and an M.S. degree
in 2017 from the Ulsan National Institute of Science and Technology,
Ulsan, South Korea, where he is currently pursuing a doctor's degree
in electrical and computer engineering. His research interests include
image processing, human perception, and human recognition.
\end{IEEEbiographynophoto}

%\begin{IEEEbiography}[{\includegraphics[width=1in,height=1.25in,clip,keepaspectratio]{./Biography/jangjh.pdf}}]{Jinhyeok Jang}
\begin{IEEEbiographynophoto}{Jinhyeok Jang} received a  B.S. degree in 2014 and an M.S. degree in 2016 from the
School of Electrical and Computer Engineering of the Ulsan National Institute of Science and Technology, Ulsan, South Korea. He currently works at the Electronics and Telecommunications Research Institute (ETRI), Daejeon, South Korea. His research interests include image processing, blur estimation, human facial recognition, and human action recognition.
\end{IEEEbiographynophoto}
%\end{IEEEbiography}

\begin{IEEEbiographynophoto}{Chanhyeok Lee} received a B.S. degree from the Ulsan National Institute of Science and Technology, Ulsan, South Korea, where he is currently pursuing a master's degree in electrical and computer engineering. His research interests include image processing, image classification, and object detection.
\end{IEEEbiographynophoto}

%\begin{IEEEbiography}[{\includegraphics[width=1in,height=1.25in,clip,keepaspectratio]{syang.pdf}}]{Seungjoon Yang } (S'99-M'00) 
\begin{IEEEbiographynophoto}{Seungjoon Yang } (S'99-M'00) 
received a B.S. degree from Seoul National University, Seoul, Korea in 1990 and M.S. and Ph.D. degrees from the University of Wisconsin-Madison in 1993 and 2000, respectively, all in electrical engineering. He worked at the Digital Media R\&D Center at Samsung Electronics Co., Ltd. from September 2000 to August 2008 and is currently with the School of Electrical and Computer Engineering at the Ulsan National Institute of Science and Technology in Ulsan, Korea. His research interests include image processing, estimation theory, and multi-rate systems.
Professor Yang received the Samsung Award for the Best Technology Achievement of the Year in 2008 for his work on  the premium digital television platform project.
\end{IEEEbiographynophoto}

\vfill

\end{document}